\documentclass[letterpaper,journal]{IEEEtran}

\usepackage{amsmath,amsfonts}
\usepackage{algorithmic}
\usepackage{algorithm}
\usepackage{array}
\usepackage[caption=false,font=normalsize,labelfont=sf,textfont=sf]{subfig}
\usepackage{textcomp}
\usepackage{stfloats}
\usepackage{url}
\usepackage{verbatim}
\usepackage{graphicx}
\usepackage{microtype}
\usepackage{booktabs}
\usepackage{adjustbox}
\usepackage{makecell}
\usepackage{multirow}
\usepackage{pifont}
\usepackage[table]{xcolor}
\usepackage{threeparttable}
\usepackage{tabularx}
\usepackage{placeins}
\usepackage[colorlinks=true, citecolor=blue, linkcolor=blue, urlcolor=black]{hyperref}
\usepackage{silence}
\makeatletter
\def\@cite#1#2{{\color{blue}[{#1\if@tempswa , #2\fi}]}}
\makeatother
\newcommand{\cmark}{\ding{51}} % ✓
\newcommand{\xmark}{\ding{55}} % ✗
\newcommand{\corresponding}{\textsuperscript{\dag}}
\definecolor{best}{RGB}{255,0,0}     % red   - 1st best
\definecolor{second}{RGB}{0,0,255}   % blue  - 2nd best
\definecolor{third}{RGB}{0,220,20}   % green - 3rd best
\begin{document}

\title{CC-Pan: Channel-wise Compression based Diffusion for Efficient Pan-Sharpening}

\author{
Junjie~Li,
Congyang~Ou,
Haokui~Zhang\corresponding,
Guoting~Wei,
Shengqin~Jiang,
and~Ying~Li
% \thanks{This work was supported in part by XXX under Grant XXX.}
\thanks{Junjie Li, Congyang Ou, Haokui Zhang, and Ying Li are with the School of Cyberspace Security, Northwestern Polytechnical University, Xi'an, Shanxi, China (e-mail: hkzhang@nwpu.edu.cn).}
\thanks{Guoting Wei is with the Nanjing University of Science and Technology, Nanjing, Jiangsu, China.}
\thanks{Shengqin Jiang is with the School of Computer Science, Nanjing University of Information Science and Technology, Nanjing, Jiangsu, China.}
\thanks{Corresponding author: Haokui Zhang (e-mail: hkzhang@nwpu.edu.cn).}
\thanks{Our code is available at \url{https://github.com/JJLibra/CC-Pan}.}
}

\maketitle

\begin{abstract}
Recently, diffusion models have brought novel insights to pan-sharpening
and notably boosted fusion precision. However, most existing models
perform diffusion in the pixel space and train distinct models for
different multispectral (MS) sensors, suffering from high inference
latency and sensor-specific limitations. In this paper, we present
CC-Pan, a cross-sensor latent diffusion framework for efficient
pan-sharpening. Specifically, CC-Pan trains a band-wise
single-channel variational autoencoder (VAE) to encode
high-resolution multispectral (HRMS) images into compact latent
representations, naturally supporting MS images with varying band
counts across different sensors and establishing a basis for
inference acceleration. Spectral physical properties, along with PAN
and MS images, are then injected into the diffusion backbone through
carefully designed unidirectional and bidirectional interactive
control structures, achieving high-precision spatial--spectral fusion
in the latent diffusion process. Furthermore, a lightweight
region-based cross-band attention (RCBA) module is incorporated at
the central layer of the diffusion model, reinforcing inter-band
spectral connections to boost spectral consistency and further
elevate fusion precision. Extensive experimental results on
GaoFen-2, QuickBird, and WorldView-3 demonstrate that CC-Pan
outperforms state-of-the-art diffusion-based methods across all
three benchmarks, attains a $2$--$3\times$ inference speedup, and
exhibits robust cross-sensor generalization capability on the
held-out WorldView-2 sensor without any sensor-specific retraining.
\end{abstract}


\section{Introduction}
\label{sec:intro}

\IEEEPARstart{E}{xisting} earth observation systems face a fundamental trade-off between spatial and spectral resolution due to physical sensor limitations~\cite{review1_2023}. These systems can only capture either low-resolution multispectral (LRMS) images with rich spectral information or high-resolution panchromatic (PAN) images with a single band. Pansharpening addresses this limitation by fusing LRMS and PAN images to generate high-resolution multispectral (HRMS) imagery that possesses both high spatial resolution and rich spectral information, which is critical for land cover mapping~\cite{land_cover_2023}, change detection~\cite{urban_change_detection_2025}, and disaster assessment~\cite{damage_assessment_2024}.

Traditional pansharpening methods typically include component substitution (CS) methods~\cite{cs_2014}, multi-resolution analysis (MRA) methods~\cite{mra_2002}, and variational optimization (VO) techniques~\cite{vo_2017}. These approaches employ hand-crafted transformations, multi-scale decompositions, or optimization models. However, they struggle to optimally balance spatial enhancement and spectral preservation, often yielding suboptimal results~\cite{cs_mra_vo_review_2021}. In recent years, deep learning methods have gained prominence by learning complex nonlinear mappings directly from data. Convolutional neural networks (CNNs) and Transformer-based pansharpening methods, leveraging their strong local and global modeling capabilities, have achieved superior performance over traditional methods~\cite{cnn_trans_2024}. Despite this progress, effectively capturing the complex cross-modal joint distribution between PAN and MS modalities to generate high-quality HRMS with both rich spectral and spatial information remains a fundamental challenge.

Diffusion models have recently emerged as a promising alternative for pansharpening~\cite{ddif_2024,taming_2025}. By modeling complex distributions through iterative denoising processes, they can effectively learn cross-modal relationships between PAN and MS modalities~\cite{ddrm_2022}. PanDiff~\cite{pandiff2023} first introduces denoising diffusion probabilistic models (DDPM)~\cite{ddpm_2020} to pansharpening, where PAN and LRMS images serve as conditional inputs to guide the denoising process, rather than being directly fused as in aforementioned methods. Subsequent works have further advanced this paradigm by refining the conditioning mechanisms, such as introducing spatial-spectral decomposition~\cite{ssdiff2024} to separately model spatial and spectral features, and employing semantic-guided routing~\cite{sgdiff2025} to enhance cross-sensor generalization. Despite these advances, existing diffusion-based pansharpening methods still face two critical limitations: 1) they typically perform diffusion directly at the pixel level, resulting in substantial computational overhead and high inference latency; 2) they are often sensor-specific, requiring distinct models to be trained for different satellite sensors due to variations in spectral band configurations across sensors.

To address these challenges, we propose \textit{CC-Pan}, a cross-sensor latent diffusion framework for efficient pansharpening. \textit{CC-Pan} employs a two-stage training strategy. In Stage I, we train a band-wise single-channel VAE to encode each spectral band of HRMS independently into a compact latent representation. This band-wise encoding strategy enables handling arbitrary spectral bands across different sensors, establishing a solid foundation for both computational efficiency and cross-sensor generalization.
In Stage II, we perform conditional latent diffusion with disentangled spatial-spectral conditioning. HRMS is first encoded band-by-band into latent features using the pretrained VAE encoder. We then inject multi-modal conditioning through carefully designed control structures: PAN provides spatial guidance, LRMS provides spectral information, and frozen CLIP~\cite{clip_2021} text embeddings provide sensor-aware physical metadata, enabling a single unified model to adapt to different sensors without retraining.
Notably, we design bidirectional interactions between both conditioning branches (spatial and spectral) and the diffusion backbone, allowing adaptive feature refinement throughout the denoising process. This disentangled yet interactive strategy effectively captures the cross-modal joint distribution between PAN and MS modalities, enabling high-precision fusion in the compact latent space.
Furthermore, to address the potential band independence introduced by the band-wise VAE processing, we introduce a lightweight Region-based Cross-Band Attention (RCBA) module that captures dependencies across spectral bands, thereby enhancing spectral consistency and further improving fusion quality. Our main contributions are:

\begin{itemize}
    \item We propose \textit{CC-Pan}, a cross-sensor latent diffusion framework for pansharpening. A band-wise single-channel VAE is trained to map each spectral band independently into a unified compact latent space, enabling both inference acceleration and sensor-agnostic band handling. To the best of our knowledge, \textit{CC-Pan} is the first pansharpening method to perform latent diffusion in a latent space that generalizes across sensors with varying spectral band counts without retraining.
    \item A bidirectional interaction design is proposed to guide the diffusion process, where PAN, LRMS, and sensor-specific prompts are jointly injected. More fine-grained interactions better coordinate the conditioning signals, and iterative feature refinement further improves performance. Additionally, a lightweight RCBA module is introduced to enhance inter-band consistency and fusion precision.
    \item Extensive experiments demonstrate that \textit{CC-Pan} outperforms previous methods across three benchmarks, achieves 2--3$\times$ inference speedup, and exhibits strong cross-sensor generalization capability.
\end{itemize}

\begin{figure*}[t]
    \centering
    \includegraphics[width=0.95\textwidth]{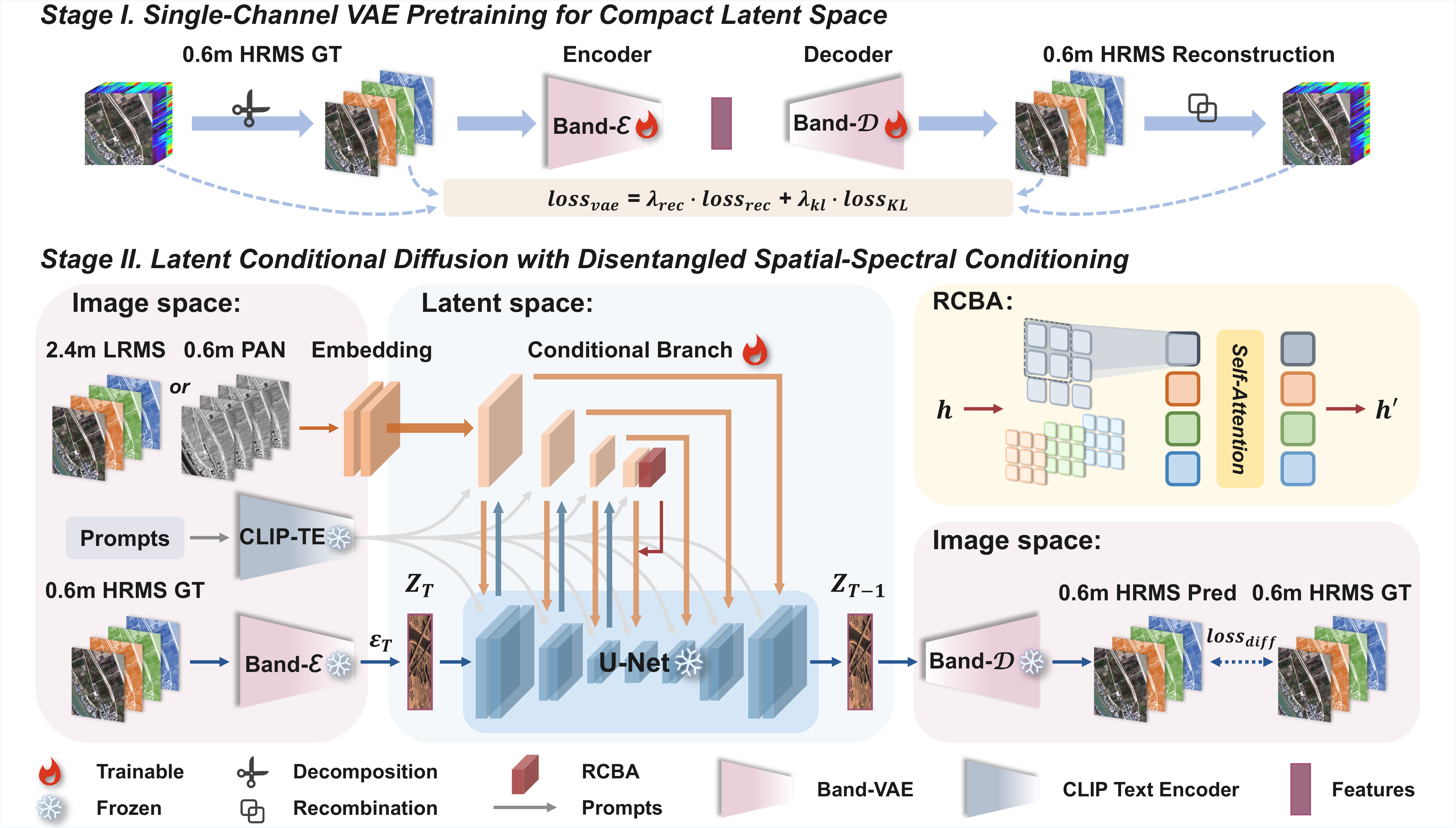}
    \caption{\textbf{Overview of \textit{CC-Pan}.}
    \textit{\textbf{Stage~I}} trains a band-wise single-channel VAE to map each HRMS band into a compact latent space.
    \textit{\textbf{Stage~II}} performs band-wise conditional latent diffusion with disentangled spatial--spectral conditioning: a spatial branch encodes PAN for spatial guidance, while a spectral branch encodes the upsampled LRMS band-by-band for spectral guidance.
    We use hybrid coupling: bidirectional interaction in the encoder and unidirectional (branch$\rightarrow$backbone) control in the mid block and decoder.
    RCBA improves inter-band consistency, and sensor-aware metadata prompts from a frozen CLIP text encoder provide additional conditioning.}
    \label{fig:cloud_overview}
\end{figure*}

\section{Related Work}
\label{sec:related_work}

\textit{\textbf{Traditional Methods.}}
Conventional pansharpening approaches are broadly categorized into three families: component substitution (CS), multi-resolution analysis (MRA), and variational optimization (VO)~\cite{cs_mra_vo_review_2021}.
CS methods project MS imagery into alternative color spaces (e.g., IHS, PCA, GS), substitute the spatial component with PAN information, and invert the transform. While computationally efficient, these methods frequently introduce spectral distortion~\cite{cs_2014}.
MRA methods inject high-frequency spatial details derived from PAN into upsampled MS through multi-scale decompositions, such as wavelet or Laplacian pyramid transforms, achieving better spectral preservation at the cost of spatial artifacts~\cite{mra_2002}.
VO methods cast pansharpening as a regularized energy minimization problem, offering theoretical guarantees but at the expense of high computational complexity and sensitivity to hyperparameter tuning~\cite{vo_2017}.
Despite their diversity, all three families rely on hand-crafted priors and struggle to optimally balance spatial enhancement with spectral preservation.

\textit{\textbf{Deep Learning Approaches.}}
Data-driven methods have come to dominate pansharpening by learning complex nonlinear spatial-spectral mappings directly from imagery~\cite{dl_benchmark_2022,full_res_2022,adwm_2025}.
CNN-based approaches, such as PNN~\cite{pnn_2016} and PanNet~\cite{pannet2017}, employ multi-scale convolutional architectures to extract local spatial-spectral features from concatenated PAN and MS inputs.
Transformer-based methods, including PanFormer~\cite{panformer_2022} and PanAdapter~\cite{panadapter_2025}, leverage long-range self-attention to capture global spatial-spectral dependencies, achieving superior fusion quality.
However, both paradigms adopt deterministic regression frameworks that map inputs directly to HRMS predictions, without explicitly modeling the underlying cross-modal joint distribution between PAN and MS modalities~\cite{ddrm_2022}. Furthermore, these methods are typically trained for a specific sensor, limiting their applicability across different satellite platforms.

\textit{\textbf{Diffusion Models.}}
Diffusion-based generative models have recently emerged as a compelling paradigm for pansharpening, offering a principled approach to capturing the complex PAN--MS cross-modal distribution through iterative denoising~\cite{crossdiff_2024,tmdiff_2024,frdiff_2025,uk_diffpan_2025}.
PanDiff~\cite{pandiff2023} pioneers this direction by adapting DDPM~\cite{ddpm_2020} to pansharpening, using jointly concatenated PAN and LRMS as conditional inputs to guide the denoising process.
SSDiff~\cite{ssdiff2024} further refines the conditioning strategy through spatial--spectral decomposition to separately model spatial and spectral components, while SGDiff~\cite{sgdiff2025} employs semantic-guided routing to improve cross-sensor feature alignment.
Despite these advances, existing diffusion-based methods share two fundamental limitations that \textit{CC-Pan} is designed to address: (i)~iterative denoising in the full-resolution pixel space incurs substantial computational overhead, as each denoising step operates on the entire image; and (ii)~fixed-dimensional band encodings couple model architectures to specific sensor configurations, necessitating retraining whenever the spectral band count changes.

\section{Methodology}
\label{sec:method}

\subsection{Problem Formulation}
Pansharpening fuses a low-resolution multispectral (LRMS) image $\mathbf{M}\in\mathbb{R}^{B\times h\times w}$ and a high-resolution panchromatic (PAN) image $\mathbf{P}\in\mathbb{R}^{1\times H\times W}$ to generate a high-resolution multispectral (HRMS) image $\mathbf{X}\in\mathbb{R}^{B\times H\times W}$, where $B$ denotes the number of spectral bands and the spatial scale ratio is $r = H/h = W/w$. The objective is to reconstruct $\mathbf{X}$ that preserves the spectral characteristics of $\mathbf{M}$ while capturing the spatial details from $\mathbf{P}$.

Existing diffusion-based pansharpening methods perform denoising directly in pixel space on full-resolution HRMS images throughout the diffusion chain, resulting in high computational cost and slow inference. In contrast, \textit{CC-Pan} performs diffusion in a compact latent space with band-wise processing, where each band $\mathbf{X}^{(b)}\in\mathbb{R}^{1\times H\times W}$ ($b=1,\ldots,B$) is encoded and processed independently, substantially reducing computational overhead.

\subsection{Framework Overview}
\label{sec:overview}
As illustrated in Figure~\ref{fig:cloud_overview}, \textit{CC-Pan} employs a two-stage training strategy.
\textbf{\textit{Stage I: Single-Channel VAE Pretraining.} }
We train a band-wise single-channel VAE that encodes each spectral band of HRMS independently into compact latent representations. This band-wise processing strategy naturally supports arbitrary numbers of spectral bands, enabling cross-sensor generalization.
\textbf{\textit{Stage II: Latent Conditional Diffusion.} }
With the VAE encoder frozen, we perform conditional diffusion in the latent space. The diffusion backbone receives disentangled spatial-spectral conditioning from PAN and LRMS images via dual control branches, along with sensor-aware physical metadata via text cross-attention. A lightweight region-based cross-band attention (RCBA) module at the central layer further enhances spectral consistency.

\subsection{Band-wise Single-channel VAE}
\label{sec:vae}

In Stage I, we train a VAE to enable diffusion in a compact latent space, which significantly reduces computational cost compared to pixel-space diffusion. However, existing VAEs (e.g., SD-VAE~\cite{sd_v15_2022}) are typically designed for three-channel RGB images, creating an application gap for HRMS images with varying numbers of spectral bands across different sensors. To bridge this gap, we propose a band-wise single-channel VAE that processes each spectral band independently through a shared encoder-decoder architecture. This design naturally supports arbitrary band configurations across sensors and avoids inter-band interference, enabling precise encoding of each band's spectral characteristics.

Specifically, we modify the first and last convolutional layers that interface with the image channels to operate on single-channel inputs. The VAE first encodes each band of HRMS $\mathbf{X}^{(b)}\in\mathbb{R}^{1\times H\times W}$ and produces a diagonal Gaussian posterior $q_{\phi}(\mathbf{z}\mid \mathbf{X}^{(b)})=\mathcal{N}(\boldsymbol{\mu}_{\phi}(\mathbf{X}^{(b)}), \mathrm{diag}(\boldsymbol{\sigma}^2_{\phi}(\mathbf{X}^{(b)})))$, from which a latent representation $\mathbf{z}\in\mathbb{R}^{C\times h'\times w'}$ is sampled, where $h'\ll H$ and $w'\ll W$. Subsequently, the decoder reconstructs the band from $\mathbf{z}$ as $\hat{\mathbf{X}}^{(b)}=\mathcal{D}_{\psi}(\mathbf{z})$.

After VAE pretraining, we use the mean of the posterior distribution as the deterministic latent representation and apply a fixed scaling factor $\kappa_{\text{vae}}$ to normalize the latent magnitude for the subsequent diffusion process:
\begin{equation}
\mathbf{z}^{(b)}_0 = \kappa_{\text{vae}}\cdot \boldsymbol{\mu}_{\phi}(\mathbf{X}^{(b)}),
\qquad
\hat{\mathbf{X}}^{(b)}=\mathcal{D}_{\psi}\!\left(\mathbf{z}^{(b)}_0/\kappa_{\text{vae}}\right),
\label{eq:vae_scale}
\end{equation}
where $\kappa_{\text{vae}}$ is estimated from empirical latent statistics and remains fixed thereafter.

\subsection{Latent Conditional Diffusion Model}
\label{sec:latent_diffusion}
In Stage~II, we train a conditional diffusion model to recover the clean latent $\mathbf{z}^{(b)}$ for each band with the pretrained VAE frozen.
The generation is conditioned on three complementary sources: (i)~the PAN image providing high-frequency spatial details, (ii)~the bicubic-upsampled LRMS band offering spectral radiometric guidance, and (iii)~sensor-aware metadata encoded as text prompts.

\paragraph{Latent diffusion formulation.}
For each band $b$, we apply a standard DDPM~\cite{ddpm_2020} forward process in the latent space:
\begin{equation}
\mathbf{z}^{(b)}_t=\sqrt{\bar{\alpha}_t}\mathbf{z}^{(b)}_0+\sqrt{1-\bar{\alpha}_t}\boldsymbol{\epsilon}^{(b)},
\qquad
\boldsymbol{\epsilon}^{(b)}\sim\mathcal{N}(\mathbf{0},\mathbf{I}),
\label{eq:ddpm_forward_latent}
\end{equation}
where $\bar{\alpha}_t=\prod_{i=1}^{t}\alpha_i$ and $t\in\{1,\ldots,T\}$.
The denoiser predicts the added noise:
\begin{equation}
\hat{\boldsymbol{\epsilon}}^{(b)}=
f_{\theta}\!\left(\mathbf{z}^{(b)}_t,t;\mathbf{P},\tilde{\mathbf{M}}^{(b)},\mathcal{E}(s_{S,b})\right),
\label{eq:eps_pred_latent}
\end{equation}
where $\tilde{\mathbf{M}}=\mathrm{Up}(\mathbf{M})\in\mathbb{R}^{B\times H\times W}$ denotes the bicubic-upsampled LRMS,
and $\mathcal{E}(s_{S,b})$ is the prompt embedding of the sensor/band descriptor $s_{S,b}$.
Given $\hat{\boldsymbol{\epsilon}}^{(b)}$, the clean-latent estimate is recovered as
\begin{equation}
\hat{\mathbf{z}}^{(b)}_0
=
\frac{\mathbf{z}^{(b)}_t-\sqrt{1-\bar{\alpha}_t}\,\hat{\boldsymbol{\epsilon}}^{(b)}}{\sqrt{\bar{\alpha}_t}}.
\label{eq:eps_to_z0}
\end{equation}

\begin{figure}[t]
    \centering
    \includegraphics[width=0.98\linewidth]{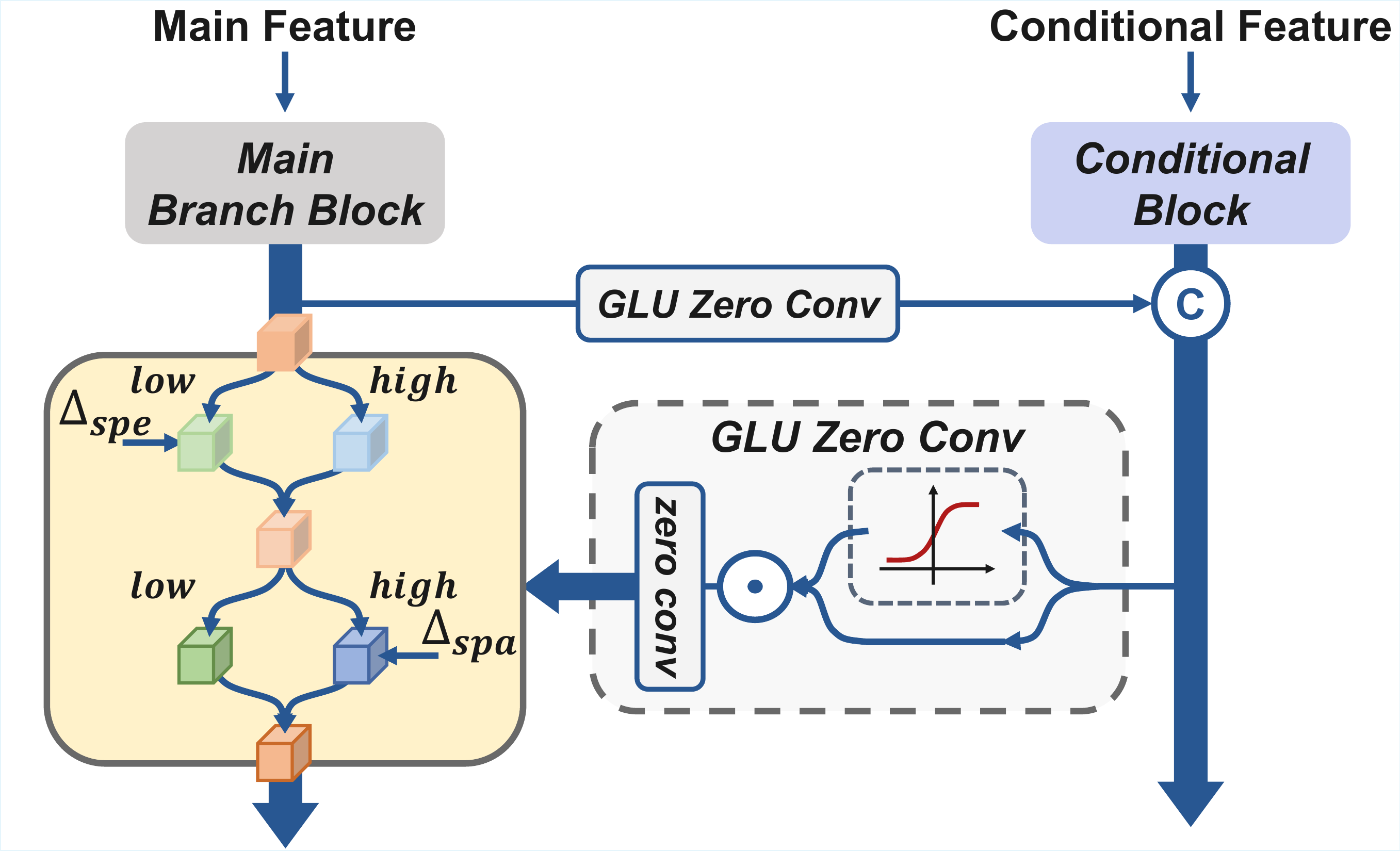}
    \caption{At each resolution, a PAN-driven spatial control branch and an LRMS-driven spectral control branch couple with the main trunk via GLU zero-convolution residual adapters: bidirectional in the encoder only, and branch$\rightarrow$trunk in the mid/decoder blocks. Branch residuals are fused via frequency-split injection (Sec.~\ref{sec:hybrid_coupling_f_split}).}
    \label{fig:condition_branch}
\end{figure}

\subsubsection{Conditional Control Mechanism}
\label{sec:hybrid_coupling}

As shown in Figure.~\ref{fig:condition_branch}, we attach two lightweight control branches to a largely frozen SD-UNet trunk:
a spatial branch conditioned on PAN $\mathbf{P}$ and a spectral branch conditioned on $\tilde{\mathbf{M}}^{(b)}$.
Each branch constructs a multi-resolution feature pyramid,
$\{\mathbf{s}^{\ell}_{\text{spa}}\}_{\ell}$ and $\{\mathbf{s}^{\ell}_{\text{spe}}\}_{\ell}$, aligned with the trunk resolutions.
Let $\mathbf{h}^{\ell}$ denote the trunk feature at resolution level $\ell$.

\paragraph{Hybrid coupling.}
At each resolution level $\ell$, the trunk $\mathbf{h}^\ell$ and the branches $\mathbf{s}^\ell_j$ ($j\in\{\text{spa},\text{spe}\}$) interact through lightweight residual adapters.
Bidirectional trunk$\leftrightarrow$branch interaction is used \emph{only in the encoder}: the trunk first feeds back to each branch, which then returns a residual $\Delta^\ell_j$ to the trunk.
Concretely, for $\ell\in\mathcal{L}_{\mathrm{enc}}$:
\begin{equation}
\tilde{\mathbf{s}}^{\ell}_{j}
=
\Phi^{\ell}_{j}\!\left([\mathbf{s}^{\ell}_{j},\ \Pi^{\ell}_{t\rightarrow j}(\mathbf{h}^{\ell})]\right),
\qquad
\Delta^{\ell}_{j}
=
\Pi^{\ell}_{j\rightarrow t}(\tilde{\mathbf{s}}^{\ell}_{j}),
\label{eq:hybrid_coupling_fb}
\end{equation}
where $[\cdot,\cdot]$ denotes channel concatenation, $\Phi^{\ell}_{j}$ is a narrow residual stack, and $\Pi^{\ell}_{t\rightarrow j}$, $\Pi^{\ell}_{j\rightarrow t}$ are lightweight residual adapters.

In the \emph{mid} and \emph{decoder} blocks, trunk$\rightarrow$branch feedback is disabled and only \emph{unidirectional} branch$\rightarrow$trunk control is retained:
\begin{equation}
\tilde{\mathbf{s}}^{\ell}_{j}
=
\Phi^{\ell}_{j}\!\left(\mathbf{s}^{\ell}_{j}\right),
\qquad
\Delta^{\ell}_{j}
=
\Pi^{\ell}_{j\rightarrow t}(\tilde{\mathbf{s}}^{\ell}_{j}),
\label{eq:hybrid_coupling_ff}
\end{equation}
for all $\ell\in\mathcal{L}_{\mathrm{mid+dec}}$, where $\mathcal{L}_{\mathrm{enc}}$ and $\mathcal{L}_{\mathrm{mid+dec}}$ denote the encoder and mid+decoder block sets, respectively.

To preserve the pretrained feature hierarchy, all coupling adapters are initialized to output zeros:
\begin{equation}
\Pi(\mathbf{x})
=
\mathrm{ZeroConv}_{1\times1}\!\big(\mathrm{GLU}(\mathrm{Conv}_{1\times1}(\mathbf{x}))\big),
\label{eq:adapter_param}
\end{equation}
where $\mathrm{ZeroConv}_{1\times1}$ is a $1{\times}1$ convolution with zero-initialized weights, and GLU denotes the gated linear unit~\cite{glu_2017}: given $\mathbf{u}=[\mathbf{u}_1,\mathbf{u}_2]$, $\mathrm{GLU}(\mathbf{u})=\mathbf{u}_1\odot\sigma(\mathbf{u}_2)$.
This guarantees $\Pi(\mathbf{x})=\mathbf{0}$ at initialization, leaving the pretrained trunk unchanged and enabling stable, gradual learning of task-specific control signals.

\paragraph{Frequency-split injection.}
\label{sec:hybrid_coupling_f_split}
PAN primarily encodes sharp edges and fine spatial textures, while LRMS mainly constrains slowly-varying spectral radiometry~\cite{freq_2020}.
To exploit this complementarity, we fuse the branch-to-trunk residuals in the frequency domain: \emph{low-frequency} components from the spectral branch and \emph{high-frequency} components from the spatial branch are injected separately:
\begin{align}
\mathbf{h}_1^{\ell} &= \mathbf{h}^{\ell}+\lambda_{\text{spe}}\mathcal{L}(\Delta^{\ell}_{\text{spe}}), \\
\mathbf{h}^{\ell}_{\text{out}} &= \mathbf{h}_1^{\ell}+\lambda_{\text{spa}}\mathcal{H}(\Delta^{\ell}_{\text{spa}}),
\label{eq:freq_inject_main}
\end{align}
where $\mathcal{L}(\cdot)$ is a fixed per-channel low-pass operator and $\mathcal{H}(\mathbf{x})=\mathbf{x}-\mathcal{L}(\mathbf{x})$ its high-pass complement.
$\mathcal{L}$ is implemented as a non-learned blur, so the frequency split introduces no additional trainable parameters.
We fix $(\lambda_{\text{spe}},\lambda_{\text{spa}})=(1,1)$ during training; the weights may optionally be adjusted at inference to control the spectral--spatial balance (Appendix~\ref{app:impl:stage2}).

\paragraph{Metadata conditioning.}
Satellite sensors differ substantially in their spectral characteristics, posing a challenge for cross-sensor generalization~\cite{sensors_2016}.
We address this by conditioning the denoising network on sensor-aware metadata through text prompts.
Each prompt $s_{S,b}$ encodes the sensor name, band index, and wavelength range, and is encoded by a frozen pretrained CLIP text encoder into the embedding $\mathcal{E}(s_{S,b})$, which is then injected into the UNet via standard cross-attention.
To maintain training efficiency, we keep the CLIP encoder and the pretrained UNet trunk frozen, and optimize only (i)~the cross-attention projection matrices $(W_Q, W_K, W_V)$~\cite{tune_a_video_2023} and (ii)~the newly introduced control modules (branches, adapters, and RCBA).

\subsubsection{Region-based Cross-Band Attention}
\label{sec:spe_cross_attn}
Although band-wise diffusion is computationally efficient, multispectral bands are physically correlated: for any fixed scene, band responses are jointly governed by shared surface and atmospheric factors, producing characteristic spectral signatures~\cite{hyperspectral_2021,hyperspectral_2017}.
Denoising each band independently may therefore cause inter-band inconsistency, or band drift.
To restore spectral coherence while preserving the efficiency of band-wise processing, we introduce a lightweight \emph{Region-based Cross-Band Attention (RCBA)} module inserted solely at the mid-block of the conditional branch, where it can correct global spectral correlations with minimal perturbation to the pretrained prior.

Let $\mathbf{F}\in\mathbb{R}^{(NB)\times C\times H'\times W'}$ denote the mid-block feature with the band dimension folded into the batch, where $N$, $B$, and $C$ are the number of scenes, spectral bands, and feature channels, respectively.
We first unfold it to $\bar{\mathbf{F}}\in\mathbb{R}^{N\times B\times C\times H'\times W'}$, then form per-band summary tokens via global average pooling (GAP):
\begin{equation}
\begin{aligned}
\mathbf{u}_{n,b} &= \mathrm{GAP}\!\left(\bar{\mathbf{F}}_{n,b}\right),\\
\mathbf{U}_{n} &= [\mathbf{u}_{n,1},\ldots,\mathbf{u}_{n,B}].
\end{aligned}
\label{eq:rcba_tokens}
\end{equation}
Multi-head self-attention is then applied over the band dimension to produce cross-band corrected tokens:
\begin{equation}
\mathbf{U}'_{n}=\mathrm{MHSA}\!\left(\mathrm{LN}(\mathbf{U}_{n})\right)\in\mathbb{R}^{B\times C}.
\label{eq:rcba_attn}
\end{equation}
Each corrected token is projected and broadcast back to its feature map as a residual update:
\begin{equation}
\bar{\mathbf{F}}'_{n,b}
=
\bar{\mathbf{F}}_{n,b}
+\eta\cdot \mathrm{Broadcast}\!\big(W_{\text{out}}(\mathbf{U}'_{n}[b])\big),
\label{eq:rcba_inject}
\end{equation}
where $\mathrm{Broadcast}(\cdot)$ spatially replicates a $C$-dimensional vector to $C\times H'\times W'$, $W_{\text{out}}:\mathbb{R}^{C}\!\rightarrow\mathbb{R}^{C}$ is a linear projection, and $\eta$ is a learnable scalar.
Both $\eta$ and $W_{\text{out}}$ are initialized to zero, so RCBA starts as an identity mapping that leaves pretrained trunk activations unchanged.
Finally, $\bar{\mathbf{F}}'$ is folded back to $(NB)\times C\times H'\times W'$.

\section{Experiments}
\label{sec:experiment}

\begin{table*}[t]
  \caption{Quantitative results on WorldView-3 (WV3). The 1st, 2nd best per column are in {\color{best}\textbf{red}}, {\color{second}\textbf{blue}}. Upper/lower blocks: feed-forward and diffusion-based methods.}
  \label{tab:wv3_table}
  \begin{center}
    \begin{small}
      \begin{sc}
        \setlength{\tabcolsep}{5pt}
        \renewcommand{\arraystretch}{1.15}
        \begin{adjustbox}{max width=\textwidth,center}
          \begin{tabular}{cc|cccc|ccc}
            \toprule
            \multicolumn{2}{c}{Methods} &
            \multicolumn{4}{c}{Reduced Resolution} &
            \multicolumn{3}{c}{Full Resolution} \\
            \cmidrule(lr){1-2}\cmidrule(lr){3-6}\cmidrule(lr){7-9}
            Models & Pub/Year &
            $Q_8\uparrow$ & $\text{SAM}\downarrow$ & $\text{ERGAS}\downarrow$ & $\text{SCC}\uparrow$ &
            $D_{\lambda}\downarrow$ & $D_s\downarrow$ & $\text{HQNR}\uparrow$ \\
            \midrule
            PanNet    & ICCV'17    & $0.891\!\pm\!0.045$ & $3.613\!\pm\!0.787$ & $2.664\!\pm\!0.347$ & $0.943\!\pm\!0.018$ & $0.017\!\pm\!0.008$ & $0.047\!\pm\!0.014$ & $0.937\!\pm\!0.015$ \\
            FusionNet & TGRS'20    & $0.904\!\pm\!0.092$ & $3.324\!\pm\!0.411$ & $2.465\!\pm\!0.603$ & $0.958\!\pm\!0.023$ & $0.024\!\pm\!0.011$ & $0.036\!\pm\!0.016$ & $0.940\!\pm\!0.019$ \\
            LAGConv   & AAAI'22    & $0.910\!\pm\!0.114$ & $3.104\!\pm\!1.119$ & $2.300\!\pm\!0.911$ & $0.980\!\pm\!0.043$ & $0.036\!\pm\!0.009$ & $0.032\!\pm\!0.016$ & $0.934\!\pm\!0.011$ \\
            BiMPan    & ACM MM'23  & $0.915\!\pm\!0.087$ & $2.984\!\pm\!0.601$ & $2.257\!\pm\!0.552$ & $0.984\!\pm\!0.005$ & $0.017\!\pm\!0.019$ & $0.035\!\pm\!0.015$ & $0.949\!\pm\!0.026$ \\
            ARConv    & CVPR'25    & $0.916\!\pm\!0.083$ & $2.858\!\pm\!0.590$ & $2.117\!\pm\!0.528$ & $0.989\!\pm\!0.014$ & $0.014\!\pm\!0.006$ & $0.030\!\pm\!0.007$ & $0.958\!\pm\!0.010$ \\
            WFANet    & AAAI'25    & $0.917\!\pm\!0.088$ & $2.855\!\pm\!0.618$ & $2.095\!\pm\!0.422$ & {\color{second}$\mathbf{0.989\!\pm\!0.011}$} & $0.012\!\pm\!0.007$ & $0.031\!\pm\!0.009$ & $0.957\!\pm\!0.010$ \\
            \midrule
            PanDiff   & TGRS'23    & $0.898\!\pm\!0.090$ & $3.297\!\pm\!0.235$ & $2.467\!\pm\!0.166$ & $0.980\!\pm\!0.019$ & $0.027\!\pm\!0.108$ & $0.054\!\pm\!0.047$ & $0.920\!\pm\!0.077$ \\
            SSDiff    & NeurIPS'24 & $0.915\!\pm\!0.086$ & $2.843\!\pm\!0.529$ & $2.106\!\pm\!0.416$ & $0.986\!\pm\!0.004$ & $0.013\!\pm\!0.005$ & $0.031\!\pm\!0.003$ & $0.956\!\pm\!0.016$ \\
            SGDiff    & CVPR'25    &
              {\color{second}$\mathbf{0.921\!\pm\!0.082}$} &
              {\color{second}$\mathbf{2.771\!\pm\!0.511}$} &
              {\color{second}$\mathbf{2.044\!\pm\!0.449}$} &
              $0.987\!\pm\!0.009$ &
              {\color{second}$\mathbf{0.012\!\pm\!0.005}$} &
              {\color{second}$\mathbf{0.027\!\pm\!0.003}$} &
              {\color{second}$\mathbf{0.960\!\pm\!0.006}$} \\
            \textbf{CC-Pan} & &
              {\color{best}$\mathbf{0.924\!\pm\!0.064}$} &
              {\color{best}$\mathbf{2.689\!\pm\!0.135}$} &
              {\color{best}$\mathbf{1.839\!\pm\!0.211}$} &
              {\color{best}$\mathbf{0.989\!\pm\!0.007}$} &
              {\color{best}$\mathbf{0.010\!\pm\!0.008}$} &
              {\color{best}$\mathbf{0.021\!\pm\!0.004}$} &
              {\color{best}$\mathbf{0.965\!\pm\!0.007}$} \\
            \bottomrule
          \end{tabular}
        \end{adjustbox}
      \end{sc}
    \end{small}
  \end{center}
\end{table*}

\begin{table*}[t]
  \caption{Quantitative results on QuickBird (QB). The 1st, 2nd best per column are in {\color{best}\textbf{red}}, {\color{second}\textbf{blue}}. Upper/lower blocks: feed-forward and diffusion-based methods.}
  \label{tab:qb_table}
  \begin{center}
    \begin{small}
      \begin{sc}
        \setlength{\tabcolsep}{5pt}
        \renewcommand{\arraystretch}{1.15}
        \begin{adjustbox}{max width=\textwidth,center}
          \begin{tabular}{cc|cccc|ccc}
            \toprule
            \multicolumn{2}{c}{Methods} &
            \multicolumn{4}{c}{Reduced Resolution} &
            \multicolumn{3}{c}{Full Resolution} \\
            \cmidrule(lr){1-2}\cmidrule(lr){3-6}\cmidrule(lr){7-9}
            Models & Pub/Year &
            $Q_4\uparrow$ & $\text{SAM}\downarrow$ & $\text{ERGAS}\downarrow$ & $\text{SCC}\uparrow$ &
            $D_{\lambda}\downarrow$ & $D_s\downarrow$ & $\text{HQNR}\uparrow$ \\
            \midrule
            PanNet    & ICCV'17    & $0.885\!\pm\!0.118$ & $5.791\!\pm\!0.995$ & $5.863\!\pm\!0.413$ & $0.948\!\pm\!0.021$ & $0.059\!\pm\!0.017$ & $0.061\!\pm\!0.010$ & $0.883\!\pm\!0.025$ \\
            FusionNet & TGRS'20    & $0.925\!\pm\!0.087$ & $4.923\!\pm\!0.812$ & $4.159\!\pm\!0.351$ & $0.956\!\pm\!0.018$ & $0.059\!\pm\!0.019$ & $0.052\!\pm\!0.009$ & $0.892\!\pm\!0.022$ \\
            LAGConv   & AAAI'22    & $0.916\!\pm\!0.130$ & $4.370\!\pm\!0.720$ & $3.740\!\pm\!0.290$ & $0.959\!\pm\!0.047$ & $0.085\!\pm\!0.024$ & $0.068\!\pm\!0.014$ & $0.853\!\pm\!0.018$ \\
            BiMPan    & ACM MM'23  & $0.931\!\pm\!0.091$ & $4.586\!\pm\!0.821$ & $3.840\!\pm\!0.319$ & $0.980\!\pm\!0.008$ & $0.026\!\pm\!0.020$ & $0.040\!\pm\!0.013$ & $0.935\!\pm\!0.030$ \\
            ARConv    & CVPR'25    & $0.936\!\pm\!0.088$ & $4.453\!\pm\!0.499$ & $3.649\!\pm\!0.401$ & {\color{best}$\mathbf{0.987\!\pm\!0.009}$} & {\color{second}$\mathbf{0.019\!\pm\!0.014}$} & $0.034\!\pm\!0.017$ & $0.948\!\pm\!0.042$ \\
            WFANet    & AAAI'25    & $0.935\!\pm\!0.092$ & $4.490\!\pm\!0.582$ & $3.604\!\pm\!0.337$ & {\color{second}$\mathbf{0.986\!\pm\!0.008}$} & $0.019\!\pm\!0.016$ & {\color{second}$\mathbf{0.033\!\pm\!0.019}$} & {\color{second}$\mathbf{0.948\!\pm\!0.037}$} \\
            \midrule
            PanDiff   & TGRS'23    & $0.934\!\pm\!0.095$ & $4.575\!\pm\!0.255$ & $3.742\!\pm\!0.353$ & $0.980\!\pm\!0.007$ & $0.058\!\pm\!0.015$ & $0.064\!\pm\!0.020$ & $0.881\!\pm\!0.075$ \\
            SSDiff    & NeurIPS'24 & $0.934\!\pm\!0.094$ & $4.464\!\pm\!0.747$ & $3.632\!\pm\!0.275$ & $0.982\!\pm\!0.008$ & $0.031\!\pm\!0.011$ & $0.036\!\pm\!0.013$ & $0.934\!\pm\!0.021$ \\
            SGDiff    & CVPR'25    &
              {\color{second}$\mathbf{0.938\!\pm\!0.087}$} &
              {\color{second}$\mathbf{4.353\!\pm\!0.741}$} &
              {\color{second}$\mathbf{3.578\!\pm\!0.290}$} &
              $0.983\!\pm\!0.007$ &
              $0.023\!\pm\!0.013$ &
              $0.043\!\pm\!0.012$ &
              $0.934\!\pm\!0.011$ \\
            \textbf{CC-Pan} & &
              {\color{best}$\mathbf{0.939\!\pm\!0.088}$} &
              {\color{best}$\mathbf{4.198\!\pm\!0.526}$} &
              {\color{best}$\mathbf{3.251\!\pm\!0.288}$} &
              $0.984\!\pm\!0.009$ &
              {\color{best}$\mathbf{0.017\!\pm\!0.011}$} &
              {\color{best}$\mathbf{0.026\!\pm\!0.009}$} &
              {\color{best}$\mathbf{0.957\!\pm\!0.010}$} \\
            \bottomrule
          \end{tabular}
        \end{adjustbox}
      \end{sc}
    \end{small}
  \end{center}
\end{table*}

\begin{table*}[t]
  \caption{Quantitative results on GaoFen-2 (GF2). The 1st, 2nd best per column are in {\color{best}\textbf{red}}, {\color{second}\textbf{blue}}. Upper/lower blocks: feed-forward and diffusion-based methods.}
  \label{tab:gf2_table}
  \begin{center}
    \begin{small}
      \begin{sc}
        \setlength{\tabcolsep}{5pt}
        \renewcommand{\arraystretch}{1.15}
        \begin{adjustbox}{max width=\textwidth,center}
          \begin{tabular}{cc|cccc|ccc}
            \toprule
            \multicolumn{2}{c}{Methods} &
            \multicolumn{4}{c}{Reduced Resolution} &
            \multicolumn{3}{c}{Full Resolution} \\
            \cmidrule(lr){1-2}\cmidrule(lr){3-6}\cmidrule(lr){7-9}
            Models & Pub/Year &
            $Q_4\uparrow$ & $\text{SAM}\downarrow$ & $\text{ERGAS}\downarrow$ & $\text{SCC}\uparrow$ &
            $D_{\lambda}\downarrow$ & $D_s\downarrow$ & $\text{HQNR}\uparrow$ \\
            \midrule
            PanNet    & ICCV'17    & $0.967\!\pm\!0.013$ & $0.997\!\pm\!0.022$ & $0.919\!\pm\!0.039$ & $0.973\!\pm\!0.011$ & $0.017\!\pm\!0.012$ & $0.047\!\pm\!0.012$ & $0.937\!\pm\!0.023$ \\
            FusionNet & TGRS'20    & $0.964\!\pm\!0.014$ & $0.974\!\pm\!0.035$ & $0.988\!\pm\!0.072$ & $0.971\!\pm\!0.012$ & $0.040\!\pm\!0.013$ & $0.101\!\pm\!0.014$ & $0.863\!\pm\!0.018$ \\
            LAGConv   & AAAI'22    & $0.970\!\pm\!0.011$ & $1.080\!\pm\!0.023$ & $0.910\!\pm\!0.045$ & $0.977\!\pm\!0.006$ & $0.033\!\pm\!0.013$ & $0.079\!\pm\!0.013$ & $0.891\!\pm\!0.021$ \\
            BiMPan    & ACM MM'23  & $0.965\!\pm\!0.020$ & $0.902\!\pm\!0.066$ & $0.881\!\pm\!0.058$ & $0.972\!\pm\!0.018$ & $0.032\!\pm\!0.015$ & $0.051\!\pm\!0.014$ & $0.918\!\pm\!0.019$ \\
            ARConv    & CVPR'25    & $0.982\!\pm\!0.013$ & $0.710\!\pm\!0.149$ & $0.645\!\pm\!0.127$ & {\color{second}$\mathbf{0.994\!\pm\!0.005}$} & $0.007\!\pm\!0.005$ & $0.029\!\pm\!0.019$ & $0.963\!\pm\!0.018$ \\
            WFANet    & AAAI'25    & $0.981\!\pm\!0.007$ & $0.751\!\pm\!0.082$ & $0.657\!\pm\!0.074$ & {\color{best}$\mathbf{0.994\!\pm\!0.002}$} & {\color{best}$\mathbf{0.003\!\pm\!0.003}$} & $0.032\!\pm\!0.021$ & {\color{second}$\mathbf{0.964\!\pm\!0.020}$} \\
            \midrule
            PanDiff   & TGRS'23    & $0.979\!\pm\!0.011$ & $0.888\!\pm\!0.037$ & $0.746\!\pm\!0.031$ & $0.988\!\pm\!0.003$ & $0.027\!\pm\!0.011$ & $0.073\!\pm\!0.013$ & $0.903\!\pm\!0.025$ \\
            SSDiff    & NeurIPS'24 & {\color{second}$\mathbf{0.983\!\pm\!0.015}$} & {\color{second}$\mathbf{0.670\!\pm\!0.124}$} & {\color{second}$\mathbf{0.604\!\pm\!0.108}$} & $0.991\!\pm\!0.006$ & $0.016\!\pm\!0.009$ & $0.027\!\pm\!0.027$ & $0.957\!\pm\!0.010$ \\
            SGDiff    & CVPR'25    & $0.980\!\pm\!0.011$ & $0.708\!\pm\!0.119$ & $0.668\!\pm\!0.094$ & $0.989\!\pm\!0.005$ & $0.020\!\pm\!0.013$ & {\color{second}$\mathbf{0.024\!\pm\!0.022}$} & $0.959\!\pm\!0.011$ \\
            \textbf{CC-Pan} & &
              {\color{best}$\mathbf{0.983\!\pm\!0.007}$} &
              {\color{best}$\mathbf{0.667\!\pm\!0.051}$} &
              {\color{best}$\mathbf{0.592\!\pm\!0.088}$} &
              $0.991\!\pm\!0.003$ &
              {\color{second}$\mathbf{0.005\!\pm\!0.002}$} &
              {\color{best}$\mathbf{0.022\!\pm\!0.014}$} &
              {\color{best}$\mathbf{0.973\!\pm\!0.010}$} \\
            \bottomrule
          \end{tabular}
        \end{adjustbox}
      \end{sc}
    \end{small}
  \end{center}
\end{table*}

\begin{figure*}[t]
    \centering
    \includegraphics[width=\textwidth]{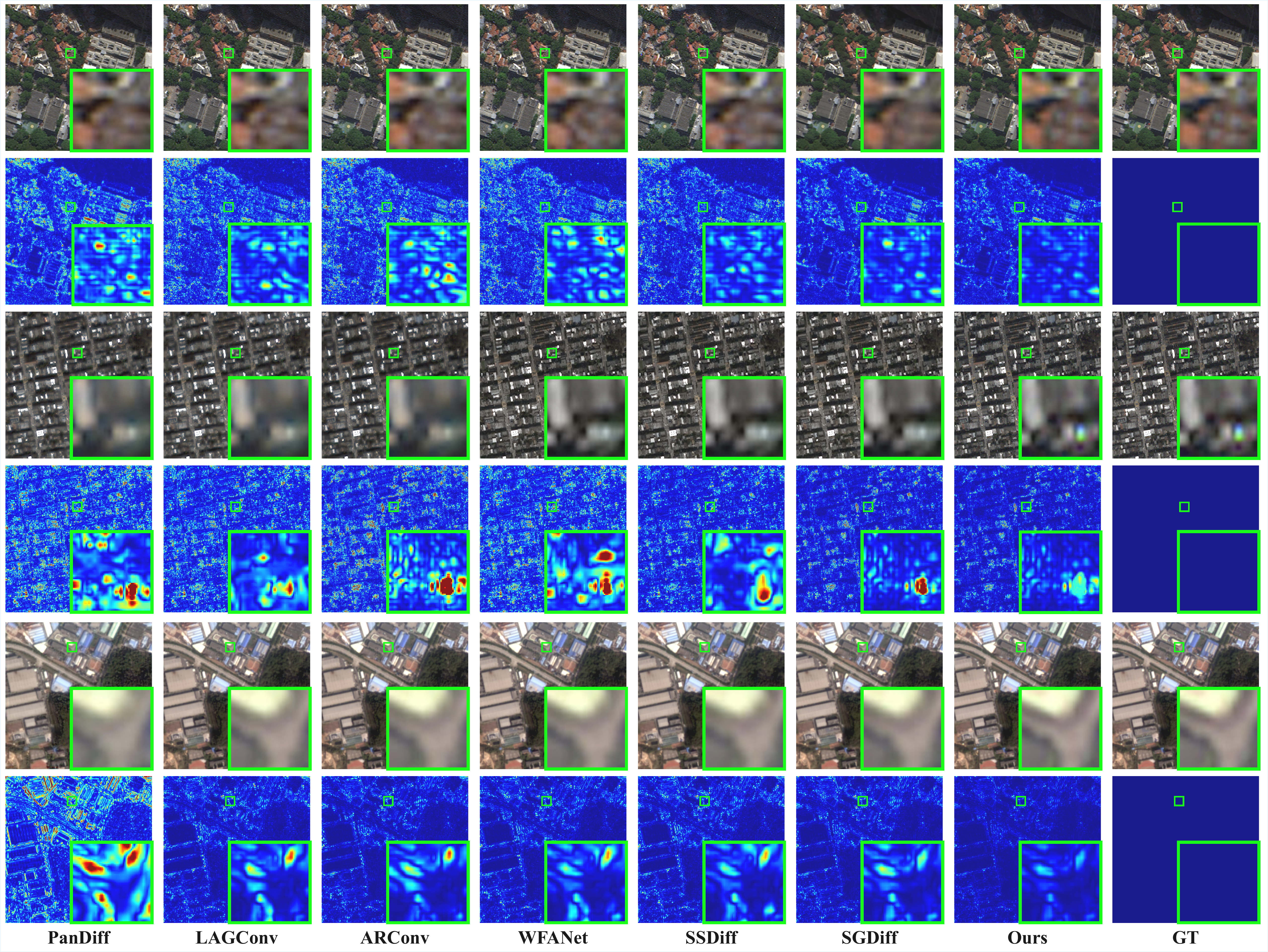}
    \caption{Visual comparison at reduced resolution (RR). Odd rows show fused images; even rows show the corresponding error heatmaps (lower error toward blue). Rows~1--2: WV3 (parking lot); Rows~3--4: QB (urban street); Rows~5--6: GF2 (industrial area). The rightmost column is the ground-truth reference. Zoomed regions (green boxes) highlight fine structures.}
    \label{fig:qual_rr_vis}
\end{figure*}

\begin{figure*}[t]
    \centering
    \includegraphics[width=\textwidth]{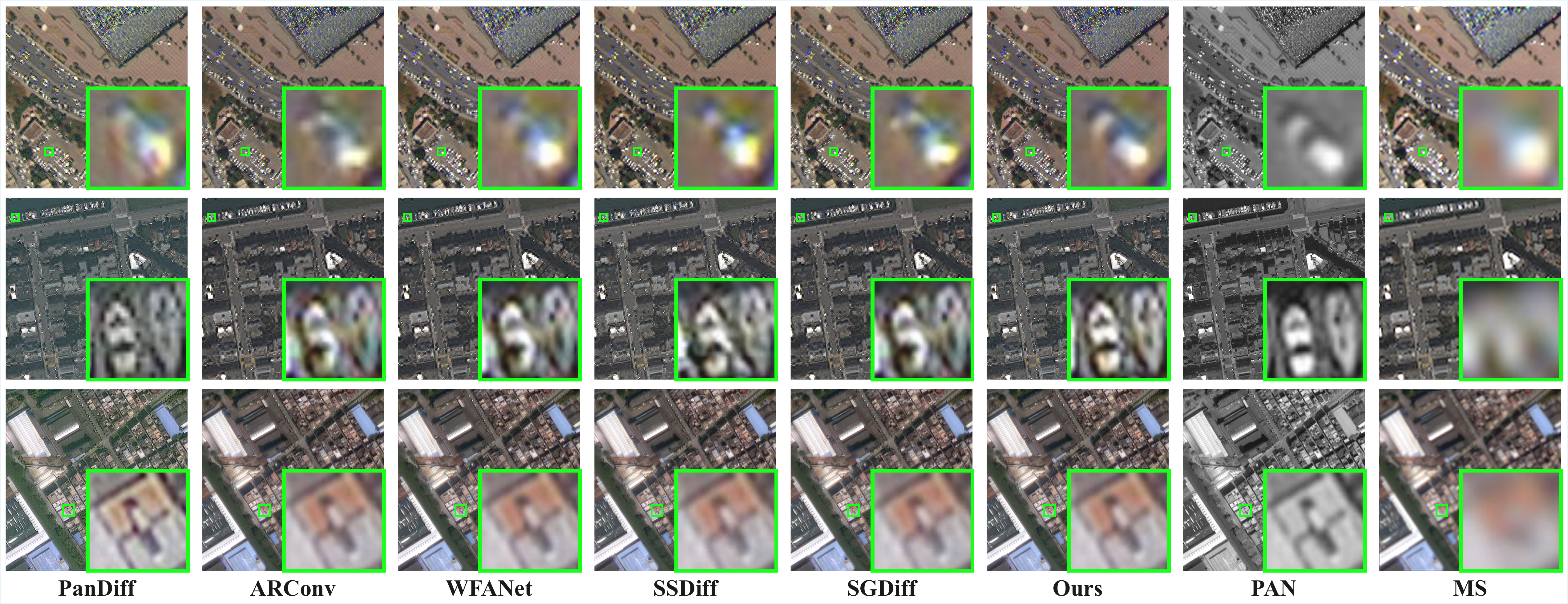}
    \caption{Visual comparison at full resolution (FR), where no ground-truth reference is available. Row~1: WV3 (parking lot); Row~2: QB (urban street); Row~3: GF2 (industrial area). Zoomed regions (green boxes) highlight spatial detail and boundary sharpness.}
    \label{fig:qual_fr_vis}
\end{figure*}

\subsection{Experimental Setup}

\paragraph{Datasets.}
We conduct experiments on the PanCollection benchmark~\cite{dl_benchmark_2022}, which provides co-registered PAN/MS image pairs from four satellite sensors: QuickBird (QB, $B{=}4$), WorldView-3 (WV3, $B{=}8$), WorldView-2 (WV2, $B{=}8$), and GaoFen-2 (GF2, $B{=}4$), all with a spatial resolution ratio of $r{=}4$.
PanCollection provides both reduced-resolution (ReducedData, RR) and full-resolution (FullData, FR) test sets with 20 samples per sensor per setting.
RR is used for reference-based evaluation under the standard Wald protocol; FR is used for real-world testing with no-reference metrics.
WV2 is held out entirely and used only to evaluate cross-sensor generalization.
We report quantitative results on WV3, QB, and GF2 as representative cases spanning both 8-band and 4-band sensors; complete results for all four sensors are provided in the supplementary material.

\paragraph{Baselines and fairness.}
We compare \textit{CC-Pan} against two groups of methods: (1)~\textbf{Feed-forward deep-learning models}: PanNet~\cite{pannet2017}, FusionNet~\cite{fusionnet2020}, LAGConv~\cite{lagconv2022}, BiMPan~\cite{bimpan2023}, ARConv~\cite{arconv2025}, and WFANet~\cite{wfanet2025}; and (2)~\textbf{Diffusion-based models}: PanDiff~\cite{pandiff2023}, SSDiff~\cite{ssdiff2024}, and SGDiff~\cite{sgdiff2025} (Appendix~\ref{app:evaluation}).
We use official implementations and released checkpoints whenever available; otherwise, baselines are retrained on the same training split with the same degradation pipeline.
All methods share identical preprocessing and postprocessing pipelines, ensuring that observed performance differences reflect model design rather than data handling.

\paragraph{Evaluation metrics.}
Under the RR Wald protocol, we report $Q_4$/$Q_8$~\cite{q4_2004,q8_2009} as overall quality indicators, SAM and ERGAS (lower is better) for spectral fidelity~\cite{sam_1992,ergas_2000}, and SCC (higher is better) for spatial-detail preservation~\cite{scc_1998}.
Under the FR setting, we follow the QNR protocol~\cite{D_s_2008,hqnr_2014,full_2022} and report $D_\lambda$ and $D_s$ (lower is better) for spectral and spatial distortion, and HQNR (higher is better) as a composite no-reference quality score.

\paragraph{Implementation details.}
\textit{CC-Pan} is implemented in PyTorch and trained for 100k steps with AdamW~\cite{AdamW_2017} (learning rate $2\times10^{-4}$, batch size $N{=}24$).
Each input batch is reshaped from $(N,B,H,W)$ to $(NB,1,H,W)$ by folding the spectral dimension into the batch dimension.
Since $B$ differs across sensors (GF2/QB: $B{=}4$; WV2/WV3: $B{=}8$), each optimization step draws from a single sensor, and a round-robin scheduler cycles through all datasets while sharing a single set of model parameters.
The Stage-I VAE is pretrained once on the three training sensors and achieves a reconstruction PSNR exceeding 60\,dB, confirming that latent compression introduces negligible information loss.
All experiments are conducted on a single RTX~4090 GPU; inference uses the UniPC sampler~\cite{unipc_2023} with $K{=}20$ steps.
All reported results use fixed test-time hyperparameters $(\lambda_\text{spe},\lambda_\text{spa}){=}(1,1)$ with no per-image or per-dataset tuning.

\subsection{Main Results}

\paragraph{Quantitative results.}
Tables~\ref{tab:wv3_table}, \ref{tab:qb_table}, and~\ref{tab:gf2_table} report results on WV3, QB, and GF2 under both RR and FR settings.
Under RR, \textit{CC-Pan} achieves the strongest spectral performance across all three sensors: on WV3, ERGAS reaches $1.839$, a substantial margin over the previous best of $2.044$, while $Q_8$ and SCC remain competitive; on GF2, \textit{CC-Pan} attains the lowest SAM ($0.667$) and ERGAS ($0.592$) among all methods.
Under FR, \textit{CC-Pan} achieves the highest HQNR on WV3 ($0.965$) and GF2 ($0.973$) together with the lowest $D_s$, confirming effective spatial distortion control under real acquisition conditions.
Since all methods share identical preprocessing and postprocessing pipelines, these gains are attributable to model design rather than data handling.

\paragraph{Qualitative comparison.}
Figures~\ref{fig:qual_rr_vis} and~\ref{fig:qual_fr_vis} show representative visual comparisons on WV3, QB, and GF2 under RR and FR, respectively.
Under RR, \textit{CC-Pan} consistently yields error heatmaps close to the ideal all-blue reference, with markedly weaker responses around boundaries and textured regions across all three sensors.
Under FR, \textit{CC-Pan} produces sharper spatial details consistent with the PAN reference while preserving the global spectral chromaticity of the MS input, avoiding the over-sharpening and local color shifts seen in competing methods.
In the WV3 example, vehicle silhouettes and boundary contrast are better preserved, whereas competing methods tend to smear fine structures or introduce visible halos.
All visualizations share identical band-to-RGB mappings and contrast stretching.

\subsection{Cross-Sensor Generalization and Efficiency}

\paragraph{Cross-sensor generalization.}
We evaluate cross-sensor capability under three transfer settings of increasing difficulty, all without target-sensor fine-tuning (Table~\ref{tab:cross_sensor}).
\textbf{Setting~1} (GF2+QB+WV3$\rightarrow$WV2, 8$\rightarrow$8 bands): a single jointly trained model is applied directly to the held-out WV2. Although WV2 and WV3 share the same 8-band configuration and originate from the same manufacturer, they differ in ground sampling distance (1.84\,m vs.\ 1.24\,m) and exhibit distinct spectral response functions, making this a non-trivial transfer.
\textbf{Setting~2} (GF2$\rightarrow$QB, 4$\rightarrow$4 bands): training on a single sensor (GF2) and testing on a different 4-band sensor from a different manufacturer, evaluating transfer across sensor origin and spatial resolution.
\textbf{Setting~3} (GF2$\rightarrow$WV2, 4$\rightarrow$8 bands): the most challenging scenario, requiring simultaneous generalization across sensor origin, spatial resolution, and spectral band count; WFANet is inapplicable here due to its fixed-dimensional band encoding.
\textit{CC-Pan} achieves the best or highly competitive performance in all three settings. The advantage is most pronounced in Setting~3, where the band-wise latent design and metadata conditioning enable generalization to an unseen band configuration, defeating fixed-architecture baselines outright.

\begin{table}[!t]
  \caption{Cross-sensor generalization on three settings of increasing difficulty (RR). No target-sensor fine-tuning is used. $^\dagger$Inapplicable due to fixed-dimensional band encoding.}
  \label{tab:cross_sensor}
  \begin{center}
    \begin{small}
      \begin{sc}
        \setlength{\tabcolsep}{4pt}
        \renewcommand{\arraystretch}{1.12}
        \begin{adjustbox}{max width=\columnwidth}
          \begin{tabular}{lcccc}
            \toprule
            \textbf{Methods} & \textbf{$Q\uparrow$} & \textbf{SAM$\downarrow$} & \textbf{ERGAS$\downarrow$} & \textbf{SCC$\uparrow$} \\
            \midrule
            \multicolumn{5}{l}{\normalfont\textit{Setting 1: GF2+QB+WV3 $\rightarrow$ WV2 (8$\rightarrow$8 bands)}} \\
            FusionNet          & $0.796\!\pm\!0.074$ & $6.426\!\pm\!0.860$ & $5.136\!\pm\!0.515$ & $0.874\!\pm\!0.013$ \\
            LAGConv            & $0.806\!\pm\!0.084$ & $6.955\!\pm\!0.474$ & $5.326\!\pm\!0.319$ & $0.912\!\pm\!0.011$ \\
            BiMPan             & $0.837\!\pm\!0.074$ & $5.328\!\pm\!0.502$ & $4.207\!\pm\!0.448$ & $0.902\!\pm\!0.016$ \\
            WFANet             & $0.845\!\pm\!0.080$ & $5.715\!\pm\!0.544$ & $4.380\!\pm\!0.466$ & $0.931\!\pm\!0.006$ \\
            SSDiff             & $0.858\!\pm\!0.078$ & $5.065\!\pm\!0.564$ & $3.989\!\pm\!0.430$ & $0.928\!\pm\!0.012$ \\
            SGDiff             & $0.858\!\pm\!0.082$ & $4.968\!\pm\!0.511$ & $\mathbf{3.868\!\pm\!0.449}$ & $0.930\!\pm\!0.007$ \\
            \textbf{CC-Pan} & $\mathbf{0.874\!\pm\!0.077}$ & $\mathbf{4.689\!\pm\!0.453}$ & $3.910\!\pm\!0.332$ & $\mathbf{0.934\!\pm\!0.009}$ \\
            \midrule
            \multicolumn{5}{l}{\normalfont\textit{Setting 2: GF2 $\rightarrow$ QB (4$\rightarrow$4 bands)}} \\
            WFANet             & $0.923$ & $4.588$ & $3.781$ & $0.970$ \\
            SGDiff             & $0.920$ & $4.597$ & $3.823$ & $0.975$ \\
            \textbf{CC-Pan} & $\mathbf{0.929}$ & $\mathbf{4.388}$ & $\mathbf{3.589}$ & $\mathbf{0.979}$ \\
            \midrule
            \multicolumn{5}{l}{\normalfont\textit{Setting 3: GF2 $\rightarrow$ WV2 (4$\rightarrow$8 bands, most challenging)}} \\
            WFANet$^\dagger$   & \multicolumn{4}{c}{Not applicable} \\
            SGDiff             & $0.842$ & $5.112$ & $4.351$ & $0.920$ \\
            \textbf{CC-Pan} & $\mathbf{0.866}$ & $\mathbf{4.720}$ & $\mathbf{4.019}$ & $\mathbf{0.925}$ \\
            \bottomrule
          \end{tabular}
        \end{adjustbox}
      \end{sc}
    \end{small}
  \end{center}
\end{table}

\paragraph{Efficiency analysis.}
We benchmark inference efficiency on a single RTX~4090 GPU with batch size one.
Wall-clock latency covers all spectral bands but excludes data loading and pre/post-processing; we also report the number of function evaluations (NFE)~\cite{nfe_2022} and per-step latency (s/step) to separately quantify the sampling budget and the per-step computational cost of the diffusion backbone.
As shown in Table~\ref{tab:efficiency}, \textit{CC-Pan} achieves the best restoration quality with only 20 NFE at $3.36$\,s per image.
On a per-step basis, \textit{CC-Pan} ($0.168$\,s/step) is substantially faster than SSDiff ($1.010$\,s/step), a direct consequence of diffusion in the compact latent space, and remains competitive with SGDiff ($0.133$\,s/step) while delivering markedly better quality at half the total latency ($3.36$\,s vs.\ $6.64$\,s).
These results confirm that the efficiency gain stems from the latent-space design itself, not from a reduced sampling budget.
We further note that SGDiff relies on GeoChat~\cite{geochat_2024} for semantic-guided routing, whose inference cost is excluded from our measurement; the reported $6.64$\,s therefore underestimates SGDiff's true deployment latency, further widening the practical efficiency advantage of \textit{CC-Pan}.

\begin{table}[t]
  \caption{Inference efficiency on QB (RR). Per-step latency isolates latent-space cost from sampling budget.}
  \label{tab:efficiency}
  \begin{center}
    \begin{small}
      \begin{sc}
        \setlength{\tabcolsep}{3pt}
        \renewcommand{\arraystretch}{1.12}
        \begin{adjustbox}{max width=\columnwidth}
          \begin{tabular}{lccccc}
            \toprule
            \textbf{Methods} & \textbf{SAM$\downarrow$} & \textbf{ERGAS$\downarrow$} & \textbf{NFE} & \textbf{Latency (s)} & \textbf{s/step} \\
            \midrule
            PanDiff            & $4.575\!\pm\!0.255$ & $3.742\!\pm\!0.353$ & 1000 & $356.63\!\pm\!1.98$ & $0.357$ \\
            SSDiff             & $4.464\!\pm\!0.747$ & $3.632\!\pm\!0.275$ & 10   & $10.10\!\pm\!0.21$  & $1.010$ \\
            SGDiff             & $4.353\!\pm\!0.741$ & $3.578\!\pm\!0.290$ & 50 & $6.64\!\pm\!0.09$ & $\mathbf{0.133}$ \\
            \textbf{CC-Pan} & $\mathbf{4.198\!\pm\!0.526}$ & $\mathbf{3.251\!\pm\!0.288}$ & 20 & $\mathbf{3.36\!\pm\!0.07}$ & $0.168$ \\
            \bottomrule
          \end{tabular}
        \end{adjustbox}
      \end{sc}
    \end{small}
  \end{center}
\end{table}

\subsection{Ablation Studies}

All ablations are conducted on WV3 under the RR protocol with identical training data and inference settings, and are organized to validate each of the three proposed contributions: (i)~the band-wise latent VAE design, (ii)~the disentangled spatial-spectral conditioning mechanism, and (iii)~the RCBA module for inter-band spectral consistency.

\paragraph{Core component analysis.}
Table~\ref{tab:ablation_component} progressively introduces each design choice from a single-branch baseline and separately compares latent-space against pixel-space diffusion.
Upgrading from single-branch (1B) to dual-branch conditioning (2B) consistently improves all metrics, confirming the benefit of disentangling PAN-driven spatial guidance from LRMS-driven spectral guidance.
Adding metadata prompts (PR) yields a further gain by injecting sensor- and band-aware context into the denoising process.
Incorporating RCBA achieves the best overall performance, underscoring the importance of explicit cross-band interaction: while the band-wise VAE improves efficiency and cross-sensor flexibility, it inherently weakens inter-band coupling in the latent space; RCBA compensates by performing sample-adaptive global cross-band correction to restore dominant spectral correlations, representing a deliberate performance-efficiency trade-off.
The lower block further confirms the value of latent-space diffusion: replacing the VAE with a pixel-space U-Net causes a substantial drop across all metrics, whereas the Stage-I VAE achieves reconstruction PSNR~$>$~60\,dB, verifying negligible compression loss.

\begin{table}[t]
  \centering
  \caption{Component ablation on WV3 (RR). Upper: progressive addition from single-branch baseline. Lower: latent-space vs.\ pixel-space diffusion.}
  \label{tab:ablation_component}
  \renewcommand{\arraystretch}{1.15}
  \setlength{\tabcolsep}{4.5pt}
  \footnotesize
  \begin{tabular}{cccc cccc}
    \toprule
    Cond. & PR & RCBA & VAE &
    $Q_8\uparrow$ & SAM$\downarrow$ & ERGAS$\downarrow$ & SCC$\uparrow$ \\
    \midrule
    1B & \xmark & \xmark & \cmark & $0.920$ & $3.057$ & $2.110$ & $0.977$ \\
    2B & \xmark & \xmark & \cmark & $0.921$ & $2.809$ & $2.045$ & $0.979$ \\
    2B & \cmark & \xmark & \cmark & $0.919$ & $2.782$ & $1.994$ & $0.983$ \\
    2B & \cmark & \cmark & \cmark &
      $\mathbf{0.924}$ & $\mathbf{2.689}$ &
      $\mathbf{1.839}$ & $\mathbf{0.989}$ \\
    \midrule
    2B & \cmark & \cmark & \xmark & $0.908$ & $3.388$ & $2.197$ & $0.966$ \\
    2B & \cmark & \cmark & \cmark &
      $\mathbf{0.924}$ & $\mathbf{2.689}$ &
      $\mathbf{1.839}$ & $\mathbf{0.989}$ \\
    \bottomrule
  \end{tabular}
\end{table}

\paragraph{Conditioning mechanism analysis.}
Table~\ref{tab:ablation_conditioning} examines two core design choices in the conditioning mechanism.
\textit{Frequency-split injection}: replacing our frequency-split design with direct residual injection consistently degrades all metrics. PAN and LRMS carry complementary frequency content — PAN primarily provides high-frequency spatial textures, while LRMS constrains low-frequency spectral radiometry — and fusing both through the same channel without separation introduces feature-space interference. Our design avoids this by routing each modality to its natural frequency band.
\textit{Bidirectional interaction}: we systematically compare eight configurations covering all combinations of bidirectional and unidirectional coupling across the encoder, mid, and decoder blocks. Encoder-bidirectional with unidirectional mid/decoder achieves the best result: early feedback from the backbone to the conditioning branches enables adaptive feature alignment, while extending bidirectional coupling to later stages yields no further benefit, suggesting conditioning information is already well-aligned at the encoder stage.

\begin{table}[t]
  \centering
  \caption{Conditioning mechanism ablation on WV3 (RR). Upper: frequency-split vs.\ direct residual. Lower: eight bidirectional interaction variants.}
  \label{tab:ablation_conditioning}
  \renewcommand{\arraystretch}{1.15}
  \setlength{\tabcolsep}{4.5pt}
  \footnotesize
  \begin{tabular}{lcccc}
    \toprule
    Setting & $Q_8\uparrow$ & SAM$\downarrow$ & ERGAS$\downarrow$ & SCC$\uparrow$ \\
    \midrule
    \multicolumn{5}{l}{\normalfont\textit{Frequency-split injection}} \\
    Direct residual              & $0.919$ & $2.780$ & $1.981$ & $0.980$ \\
    Freq.-split\,(ours)          & $\mathbf{0.924}$ & $\mathbf{2.689}$ & $\mathbf{1.839}$ & $\mathbf{0.989}$ \\
    \midrule
    \multicolumn{5}{l}{\normalfont\textit{Bidirectional interaction scheme}} \\
    All unidirectional           & $0.918$ & $3.130$ & $2.106$ & $0.980$ \\
    All bidirectional            & $0.920$ & $2.967$ & $1.967$ & $0.980$ \\
    Encoder only                 & $0.916$ & $3.124$ & $2.090$ & $0.973$ \\
    Mid only                     & $0.916$ & $3.088$ & $2.122$ & $0.969$ \\
    Decoder only                 & $0.920$ & $3.013$ & $2.020$ & $0.972$ \\
    Encoder + mid                & $0.918$ & $2.898$ & $2.093$ & $0.979$ \\
    Mid + decoder                & $0.919$ & $2.903$ & $2.102$ & $0.978$ \\
    Enc.\,bidi+mid/dec.\,uni\,(ours) & $\mathbf{0.924}$ & $\mathbf{2.689}$ & $\mathbf{1.839}$ & $\mathbf{0.989}$ \\
    \bottomrule
  \end{tabular}
\end{table}

\paragraph{Metadata encoder and robustness analysis.}
Table~\ref{tab:ablation_metadata} presents two complementary analyses of the metadata conditioning design on both the seen sensor WV3 and the held-out WV2.
\textit{Encoder comparison} (upper block, evaluated on WV2): removing metadata degrades performance to a level comparable with WFANet ($Q_8{=}0.845$, Table~\ref{tab:cross_sensor}), a sensor-specific baseline that does not generalize, confirming that sensor-aware conditioning is indispensable for cross-sensor transfer. A sensor ID embedding provides only marginal benefit; a physical MLP encoding GSD and wavelength ranges yields a more meaningful gain. The frozen CLIP encoder achieves the best result by inheriting rich pretrained semantic representations of sensor descriptors while fitting naturally into the SD-UNet cross-attention interface without additional modifications.
\textit{Metadata robustness} (lower blocks): on the seen WV3, metadata provides consistent but modest gains, as the model can partially compensate through learned sensor-specific features. On the unseen WV2, the benefit is substantially larger — removing metadata brings performance close to the WFANet baseline, while partial metadata recovers much of the gap. This graceful degradation rather than catastrophic failure supports the practical deployability of \textit{CC-Pan} under variable metadata quality.

\begin{table}[t]
  \centering
  \caption{Metadata encoder comparison and robustness analysis (RR). Upper: encoder variants on held-out WV2. Lower: robustness to metadata completeness on seen WV3 and unseen WV2.}
  \label{tab:ablation_metadata}
  \renewcommand{\arraystretch}{1.15}
  \setlength{\tabcolsep}{4.5pt}
  \footnotesize
  \begin{tabular}{lcccc}
    \toprule
    Setting & $Q_8\uparrow$ & SAM$\downarrow$ & ERGAS$\downarrow$ & SCC$\uparrow$ \\
    \midrule
    \multicolumn{5}{l}{\normalfont\textit{Metadata encoder (held-out WV2)}} \\
    None                  & $0.847$ & $5.083$ & $4.106$ & $0.927$ \\
    Sensor ID embedding   & $0.848$ & $4.985$ & $4.207$ & $0.927$ \\
    Physical MLP          & $0.857$ & $4.690$ & $4.131$ & $\mathbf{0.930}$ \\
    Frozen CLIP\,(ours)   & $\mathbf{0.874}$ & $\mathbf{4.689}$ & $\mathbf{3.910}$ & $\mathbf{0.930}$ \\
    \midrule
    \multicolumn{5}{l}{\normalfont\textit{Metadata robustness on seen sensor WV3}} \\
    None                  & $0.917$ & $2.838$ & $2.141$ & $0.977$ \\
    Partial               & $0.920$ & $2.703$ & $2.065$ & $0.982$ \\
    Full\,(ours)          & $\mathbf{0.924}$ & $\mathbf{2.689}$ & $\mathbf{1.839}$ & $\mathbf{0.989}$ \\
    \midrule
    \multicolumn{5}{l}{\normalfont\textit{Metadata robustness on unseen sensor WV2}} \\
    None                  & $0.847$ & $5.083$ & $4.106$ & $0.927$ \\
    Partial               & $0.852$ & $4.985$ & $4.003$ & $0.928$ \\
    Full\,(ours)          & $\mathbf{0.874}$ & $\mathbf{4.689}$ & $\mathbf{3.910}$ & $\mathbf{0.930}$ \\
    \bottomrule
  \end{tabular}
\end{table}

\section{Conclusion}
\label{sec:conclusion}
We have presented \textit{CC-Pan}, a cross-sensor latent diffusion framework for efficient pansharpening.
At its core, a band-wise single-channel VAE maps each spectral band independently into a compact latent space, substantially reducing computational cost relative to pixel-space diffusion and eliminating the need for sensor-specific retraining by naturally accommodating arbitrary band configurations.
Building on this foundation, a disentangled spatial-spectral conditioning mechanism with bidirectional branch interaction enables PAN, LRMS, and sensor-aware text prompts to jointly and adaptively guide the denoising process.
A lightweight Region-based Cross-Band Attention (RCBA) module further restores inter-band spectral coherence that band-wise processing inherently weakens, improving overall fusion quality.
Extensive experiments across GaoFen-2, QuickBird, and WorldView-3 validate consistent state-of-the-art performance, a $2$--$3\times$ inference speedup over pixel-space diffusion baselines, and robust cross-sensor generalization to the held-out WorldView-2 sensor.
These results establish latent diffusion with band-wise processing as a principled, efficient, and practically deployable paradigm for multi-sensor pansharpening.

\section*{Acknowledgments}
This work was supported in part by the National Natural Science Foundation of China under Grant 62401471, in part by 2024 Gusu Innovation and Entrepreneurship Leading Talents Program under Grant ZXL2024333.

\bibliographystyle{IEEEtran}
\bibliography{references}

\clearpage
\appendices

\section{Appendix Outline}
\label{app:app_outline}

This appendix complements the main paper by providing:
(i) further analysis of sensor-induced PAN--MS coupling shifts that motivate our design,
(ii) reproducible datasets, protocols, evaluation settings, and implementation details, and
(iii) additional zero-shot visual comparisons for \textsc{CC-Pan}.
It is organized as follows.

\begin{itemize}
  \item \textbf{Inductive biases for cross-sensor PAN--MS coupling (Appendix~\ref{app:priors}).}
  We analyze two sources of cross-sensor shift. The first is sensor-dependent PAN--MS spectral mixing, supported by per-sensor linear PAN surrogates in Fig.~\ref{fig:prior_pan_mixing}. The second is the RR$\rightarrow$FR mismatch in high frequencies, quantified by band-wise correlation shifts in Fig.~\ref{fig:prior2}. These findings motivate our band-wise formulation and frequency-split conditioning.

  \item \textbf{Datasets and protocols (Appendix~\ref{app:datasets}).}
  We describe the PanCollection format, sensor and band configurations, patch and image sizes, and strict train, validation, and test splits in Table~\ref{tab:pancollection_stats}. We also specify the joint-training and zero-shot settings, and the RR ReducedData and FR FullData evaluation protocols.

  \item \textbf{Evaluation setup (Appendix~\ref{app:evaluation}).}
  We summarize the compared methods and clarify the metrics used in the main paper and appendix. The reduced-resolution setting is evaluated with full-reference metrics, including $Q_4/Q_8$, SAM, ERGAS, and SCC, while the full-resolution setting is evaluated with no-reference metrics, including $D_{\lambda}$, $D_s$, and HQNR.

  \item \textbf{Implementation details (Appendix~\ref{app:impl}).}
  We give complete settings for both stages. This includes the single-band VAE conversion and latent-scale calibration, diffusion training and inference hyperparameters, test-time control, reverse-step budget selection, and metadata-prompt construction in Table~\ref{tab:prompt_params}. We also detail test-time control and sampling choices in Figs.~\ref{fig:lspe_lspa}--\ref{fig:K_sweep}.

  \item \textbf{Additional zero-shot visual results (Appendix~\ref{app:quant_vis}).}
  We provide qualitative comparisons on the unseen WorldView-2 sensor under both reduced-resolution and full-resolution settings, as shown in Figs.~\ref{fig:wv2_rr_vis}--\ref{fig:wv2_fr_vis}.
\end{itemize}

\section{Sensor-Dependent PAN--MS Coupling}
\label{app:priors}

We follow Sec.~\ref{sec:method}: pansharpening reconstructs HRMS $X$ from PAN $P$ and LRMS $M$.
Most learning-based methods are trained and evaluated under reduced-resolution RR synthesis and deployed on native full-resolution FR acquisitions where HRMS ground truth is typically unavailable.
Cross-satellite studies indicate that SRF and MTF differences, together with radiometric calibration, change the effective PAN--MS coupling across sensors, and RR degradations such as Wald-type protocols may not match the native FR imaging pipeline~\cite{dl_benchmark_2022,hetssnet_2025}.
This creates systematic RR-to-FR and cross-sensor gaps when the PAN--MS relation is treated as sensor-invariant.

Up to radiometric normalization, a sensor-indexed forward model is
\begin{equation}
\begin{aligned}
M_b &= \mathcal{D}_r\!\left(k^{(S)} * X_b\right) + \epsilon_b,\\
P &= \kappa^{(S)} *
\Big(\sum\nolimits_{b=1}^{B} w_b^{(S)} X_b\Big) + \epsilon_p,
\end{aligned}
\label{eq:img_formation_appB}
\end{equation}
where $S$ indexes the sensor, $w^{(S)}$ captures SRF-induced spectral mixing, and $k^{(S)},\kappa^{(S)}$ capture sensor-specific spatial transfer.

\subsection{Prior 1: Sensor-Dependent PAN--MS Mixing}

PAN integrates radiance over a sensor-specific spectral response and can be approximated as a sensor-dependent mixture of MS bands.
Thus, the association between $P$ and MS channels varies with $S$, and cross-sensor degradation arises when a fixed mixing is implicitly assumed~\cite{dl_benchmark_2022}.
We expose this dependence through sensor and band metadata in a fixed prompt template
\begin{equation}
s_{S,b} = s^{\text{data}}_{S} \Vert s^{\text{band}}_{S,b},
\label{eq:prompt_appB}
\end{equation}
which is embedded as $\mathcal{E}(s_{S,b})$ and used to condition band-wise denoising.

\subsubsection{Empirical evidence}
\begin{figure*}[!t]
  \centering
  \includegraphics[width=0.78\linewidth]{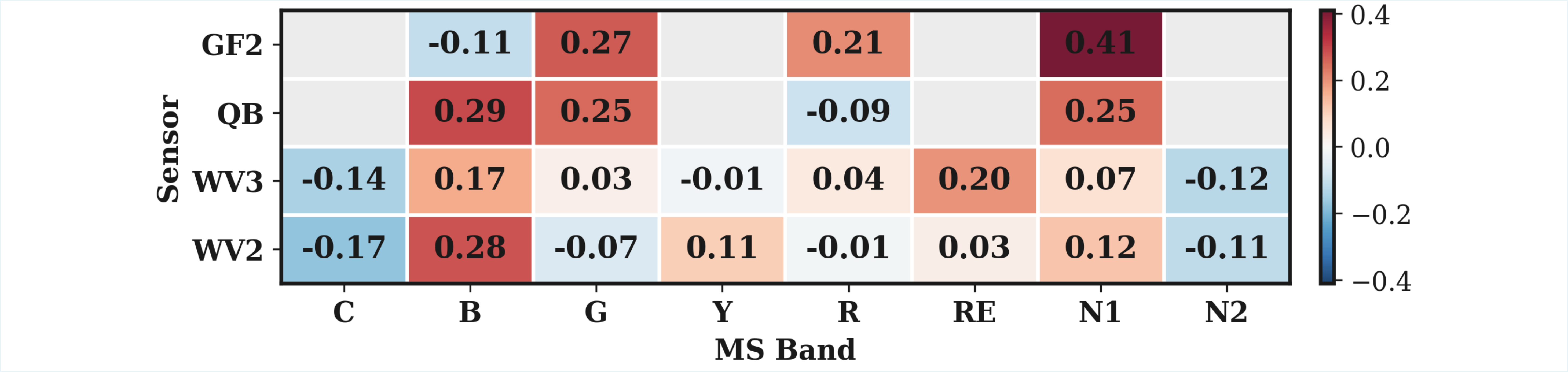}
  \caption{\textbf{Sensor-dependent PAN--MS mixing on PanCollection.}
  Heatmap of ridge-regression coefficients $\{w_b^{(S)}\}$ in Eq.~\eqref{eq:pan_mixing_surrogate_appB} fitted from aligned $\tilde M_b$ to PAN $P$.
  Rows denote sensors and columns denote MS bands $b\in\{C,B,G,Y,R,RE,N1,N2\}$.
  Blank entries indicate bands unavailable for that sensor.
  Coefficients are normalized within each sensor for visualization, and the cross-row differences reveal substantial sensor dependence in the effective PAN--MS coupling.}
  \label{fig:prior_pan_mixing}
\end{figure*}

Let $\tilde{M}=\mathcal{U}_r(M)$ denote the aligned upsampled LRMS at PAN scale, corresponding to \texttt{LMS} in PanCollection.
For each sensor $S$, we fit an effective linear surrogate
\begin{equation}
P \approx \sum_{b=1}^{B} w^{(S)}_b \tilde{M}_b + c^{(S)},
\label{eq:pan_mixing_surrogate_appB}
\end{equation}
using ridge regression on randomly subsampled pixel pairs pooled within that sensor.
Fig.~\ref{fig:prior_pan_mixing} shows substantial cross-sensor variation in $\{w_b^{(S)}\}$, which directly supports Prior~1.

\subsubsection{Connection to our formulation}
Since the coupling varies with $S$ and the number of bands varies across sensors, we avoid hard-coding a fixed multichannel interface.
We encode each band with a shared single-channel VAE,
\begin{equation}
\mathbf{z}^{(b)}_0=\kappa_{\text{vae}}\boldsymbol{\mu}_{\phi}(\mathbf{X}^{(b)}),
\label{eq:vae_scale_appB_link}
\end{equation}
and learn band-wise latent denoising conditioned on $P$, $\tilde M^{(b)}$, and metadata,
\begin{equation}
\hat{\boldsymbol{\epsilon}}^{(b)}=
f_{\theta}\!\left(\mathbf{z}^{(b)}_t,t;\mathbf{P},\tilde{\mathbf{M}}^{(b)},\mathcal{E}(s_{S,b})\right),
\label{eq:eps_pred_latent_appB_link}
\end{equation}
which explicitly represents a sensor- and band-indexed relation rather than assuming a sensor-invariant PAN--MS mixing.

\subsection{Prior 2: Sensor-Dependent Spatial Transfer Induces RR--FR Mismatch in High Frequencies}

RR synthesis replaces the native spatial transfers in Eq.~\eqref{eq:img_formation_appB} with synthetic degradations, so the PAN--MS relation learned on RR need not match FR.
The mismatch is frequency-selective and is strongest at high frequencies, which makes rigid PAN-driven detail injection brittle under RR-to-FR deployment.

Define a fixed high-pass operator $H(\cdot)=I-L(\cdot)$, where $L$ is a shared low-pass filter.
For each MS band $b$, we measure the PAN--MS high-frequency relation by
\begin{equation}
\rho_{\text{HF}}^{(b)} \;=\; \mathrm{corr}\!\left(H(P),\,H(\tilde M_b)\right),
\label{eq:hf_corr_band_appB}
\end{equation}
computed over spatial pixels per image or crop after mean subtraction.
In the Fourier domain, Eq.~\eqref{eq:img_formation_appB} implies
\begin{equation}
\begin{aligned}
\widehat{H(\tilde M_b)}(\omega)
&\approx
\widehat{H}(\omega)\,
\widehat{\mathcal{U}_r\mathcal{D}_r}(\omega)\,
\widehat{k^{(S)}}(\omega)\,
\widehat{X_b}(\omega),\\
\widehat{H(P)}(\omega)
&\approx
\widehat{H}(\omega)\,
\widehat{\kappa^{(S)}}(\omega)\,
\widehat{X_{\mathrm{mix}}^{(S)}}(\omega).
\end{aligned}
\label{eq:transfer_appB}
\end{equation}
so changes in $\widehat{k^{(S)}}$ and $\widehat{\kappa^{(S)}}$ alter relative high-frequency content and shift $\rho_{\text{HF}}^{(b)}$.

\begin{figure*}[!t]
  \centering
  \includegraphics[width=0.95\linewidth]{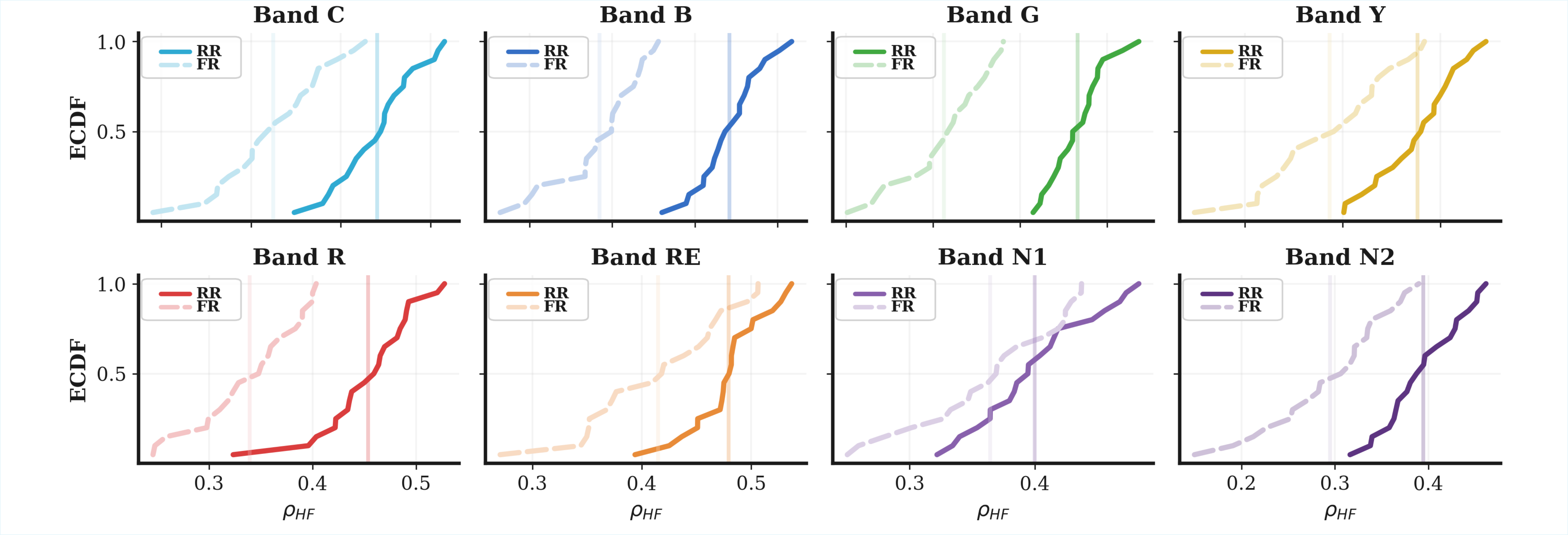}
  \caption{\textbf{WV3 RR vs. FR: PAN--MS high-frequency alignment shifts.}
  For each band $b$, we compute $\rho_{\mathrm{HF}}^{(b)}=\mathrm{corr}\!\left(H(P),\,H(\tilde M_b)\right)$ with the same $H(\cdot)$ and preprocessing in RR and FR.
  Each subplot shows the empirical CDF over images/crops.
  FR consistently shifts left, indicating weaker PAN--MS high-frequency alignment, with band-dependent magnitude; vertical lines mark means.}
  \label{fig:prior2}
\end{figure*}

We quantify the RR-to-FR shift per band by
\begin{equation}
\Delta\rho^{(b)}=\mathbb{E}_{\text{FR}}[\rho_{\text{HF}}^{(b)}]-\mathbb{E}_{\text{RR}}[\rho_{\text{HF}}^{(b)}],
\label{eq:deltaRho_band_appB}
\end{equation}
computed with identical preprocessing and sampling for RR and FR.
Fig.~\ref{fig:prior2} reports empirical CDFs of $\rho_{\text{HF}}^{(b)}$ and shows a consistent left shift from RR to FR, indicating weaker PAN--MS high-frequency alignment in native FR acquisitions and band-dependent shift magnitudes.

The RR-to-FR decrease in $\rho_{\text{HF}}^{(b)}$ indicates that PAN and band-wise $\tilde M_b$ are less aligned in high frequencies on FR, so a sensor-invariant HF injection learned on RR can over-trust PAN HF on FR.
We therefore separate spatial and radiometric guidance.
A PAN-driven spatial branch and an LRMS-driven spectral branch produce residual controls $\Delta^\ell_{\text{spa}}$ and $\Delta^\ell_{\text{spe}}$ through lightweight adapters.
These controls are fused by frequency-split injection,
\begin{equation}
\mathbf{h}^{\ell}_{\text{out}}
=
\mathbf{h}^{\ell}
+\lambda_{\text{spe}}\mathcal{L}(\Delta^{\ell}_{\text{spe}})
+\lambda_{\text{spa}}\mathcal{H}(\Delta^{\ell}_{\text{spa}}),
\qquad
\mathcal{H}(\mathbf{x})=\mathbf{x}-\mathcal{L}(\mathbf{x}),
\label{eq:freq_inject_appB_link}
\end{equation}
which anchors low-frequency radiometry to LRMS and restricts PAN guidance to high-frequency refinement, improving robustness under RR-to-FR deployment.

\section{Dataset Descriptions}
\label{app:datasets}

\subsection{Dataset Overview}

We conduct experiments on the PanCollection benchmark, which provides co-registered panchromatic and multispectral image pairs from multiple satellite sensors under two standard settings~\cite{dl_benchmark_2022}.
The reduced-resolution setting is provided as ReducedData and denoted as RR.
The full-resolution setting is provided as FullData and denoted as FR.
All sensors share the same spatial resolution ratio $r=4$ between PAN and MS.
Samples are stored in \texttt{.h5} files with layout $N\times C\times H\times W$.
Each sample includes PAN $P$ in \texttt{/PAN}, low-resolution MS $M$ in \texttt{/MS}, and the dataset-provided upsampled and co-registered MS $\tilde{M}$ at the PAN scale in \texttt{/LMS}.
RR additionally provides a pseudo ground-truth high-resolution MS target $X$ in \texttt{/GT}, enabling full-reference evaluation.
Key names may differ in letter case across releases.
We map both variants to a unified interface and use the canonical names \texttt{/PAN}, \texttt{/MS}, \texttt{/LMS}, and \texttt{/GT} throughout.

\subsection{Sensors, Band Configurations, and Patch Sizes}

PanCollection includes QuickBird, GaoFen-2, WorldView-3, and WorldView-2.
QuickBird and GaoFen-2 provide 4-band MS, while WorldView-3 and WorldView-2 provide 8-band MS.
Under RR, \texttt{/MS} has size $64\times64$ and \texttt{/PAN}, \texttt{/LMS}, \texttt{/GT} have size $256\times256$.
Under FR, \texttt{/MS} has size $128\times128$ and \texttt{/PAN} and \texttt{/LMS} have size $512\times512$.
FR does not provide \texttt{/GT}.
Table~\ref{tab:pancollection_stats} summarizes per-sensor ground sampling distance, band count, and split sizes used in this work.

\subsection{Splits, Joint Training, and Strict Zero-Shot Protocol}

We follow the official scene-level splits provided by PanCollection to prevent spatial leakage across training, validation, and testing.
In our default setting, a single model is trained on the RR training splits of QuickBird, GaoFen-2, and WorldView-3, and validated on their RR validation splits.
WorldView-2 is treated as an unseen sensor and used only for testing in both RR and FR.
No WorldView-2 samples are used for training, validation, early stopping, or any model selection.

Joint training is formulated as an equal-weight multi-sensor objective
\begin{equation}
\min_{\theta}\ \frac{1}{|\mathcal{S}_{\mathrm{train}}|}\sum_{s\in\mathcal{S}_{\mathrm{train}}}
\mathbb{E}_{(P,M,X)\sim\mathcal{D}_s}\bigl[\mathcal{L}(\theta;P,M,X,s)\bigr],
\label{eq:app_joint_train_obj}
\end{equation}
where $\mathcal{S}_{\mathrm{train}}=\{\text{QuickBird},\text{GaoFen-2},\text{WorldView-3}\}$ and $\mathcal{D}_s$ denotes the sensor-specific RR training distribution.
We implement Eq.~\eqref{eq:app_joint_train_obj} by strict round-robin scheduling at the optimizer-step level.
At step $k$, we select the sensor
\begin{equation}
s_k=\mathcal{S}_{\mathrm{train}}\bigl[k \bmod |\mathcal{S}_{\mathrm{train}}|\bigr],
\label{eq:app_round_robin}
\end{equation}
then draw a mini-batch $\mathcal{B}_k\sim\mathcal{D}_{s_k}$ and update $\theta$ using $\nabla_\theta \mathcal{L}(\theta;\mathcal{B}_k,s_k)$.
Each step uses data from a single sensor so that the MS band count is constant within the step.
The round-robin schedule yields an exact $1{:}1{:}1$ contribution over QuickBird, GaoFen-2, and WorldView-3 regardless of their dataset sizes.
All methods use the same dataset-provided \texttt{/LMS} input and the same intensity handling.

\begin{table*}[!t]
\centering
\caption{PanCollection statistics and our default usage. RR uses ReducedData and provides \texttt{/GT}. FR uses FullData and has no \texttt{/GT}. RR train and validation counts are patch-based. RR and FR test sets contain 20 images per sensor. WorldView-2 is test-only.}
\label{tab:pancollection_stats}
\small
\setlength{\tabcolsep}{4pt}
\renewcommand{\arraystretch}{1.12}
\begin{adjustbox}{max width=\linewidth,center}
\begin{tabular}{lcccccc}
\toprule
Sensor & MS/PAN GSD & Bands & RR Train patches & RR Val patches & RR Test images & FR Test images \\
\midrule
GaoFen-2   & 4.00 m / 1.00 m & 4 & 19344 & 2201 & 20 & 20 \\
QuickBird  & 2.40 m / 0.60 m & 4 & 10949 & 1905 & 20 & 20 \\
WorldView-3& 1.24 m / 0.31 m & 8 &  7787 & 1080 & 20 & 20 \\
WorldView-2& 1.84 m / 0.46 m & 8 &     0 &    0 & 20 & 20 \\
\bottomrule
\end{tabular}
\end{adjustbox}
\end{table*}

\subsection{RR Evaluation Protocol}

We use PanCollection ReducedData directly for RR evaluation.
Each RR sample provides PAN $P$ from \texttt{/PAN}, LRMS $M$ from \texttt{/MS}, aligned upsampled MS $\tilde{M}$ from \texttt{/LMS}, and pseudo ground-truth HRMS $X$ from \texttt{/GT}.
All methods take $P$ and $\tilde{M}$ as inputs and predict fused HRMS $\hat{X}$ at the PAN scale.
We report full-reference metrics computed between $\hat{X}$ and $X$.
Metric usage is summarized in Appendix~\ref{app:evaluation}.

\subsection{FR Evaluation Protocol}

We use PanCollection FullData directly for FR evaluation.
Each FR sample provides PAN $P$ from \texttt{/PAN}, LRMS $M$ from \texttt{/MS}, and aligned upsampled MS $\tilde{M}$ from \texttt{/LMS}.
All methods take $P$ and $\tilde{M}$ as inputs and predict $\hat{X}$.
Since FR does not include ground-truth HRMS, we report no-reference QNR-based measures and qualitative comparisons.
Metric usage is summarized in Appendix~\ref{app:evaluation}.

\section{Evaluation Setup}
\label{app:evaluation}

\subsection{Compared Methods}
\label{app:comp_methods}

\textbf{SGDiff}~\cite{sgdiff2025} proposes a semantic-guided diffusion model for general pansharpening across different sensors and scenes. It models the HRMS residual over the upsampled LRMS with a 3D Transformer denoiser. A pretrained geoscience VLM provides global semantics and grounded regional descriptions, which are encoded into routing scores to sparsely activate and re-weight experts in a semantic-guided MoE module.

\textbf{SSDiff}~\cite{ssdiff2024} formulates pansharpening from a subspace decomposition viewpoint, separating spatial details and spectral components within a two-branch diffusion framework. It couples the two branches via alternating projection and exchanges high-frequency cues through Fourier-based modulation during denoising. A branch-wise alternating fine-tuning strategy is further used to enhance component-specific representations.

\textbf{PanDiff}~\cite{pandiff2023} introduces a DDPM-based framework that learns the distribution of the difference map between HRMS and interpolated MS. It performs multi-step conditional denoising with a U-Net backbone, injecting PAN and LRMS as guidance at each step. A modal intercalibration module aligns spatial and spectral cues to strengthen the conditioning.

\textbf{WFANet}~\cite{wfanet2025} is a wavelet-assisted multi-frequency attention network. It decomposes PAN features into frequency bands with discrete wavelet transforms, and fuses LRMS spectra with frequency-specific spatial cues using a multi-frequency fusion attention module, followed by inverse-wavelet reconstruction. A wavelet pyramid and a spatial detail enhancement module further improve multi-scale fusion and high-frequency recovery.

\textbf{ARConv}~\cite{arconv2025} introduces an adaptive rectangular convolution that predicts kernel height and width from features and adjusts sampling accordingly. It builds a lightweight rectangular deformable sampling map and applies an affine transformation to improve spatial adaptability. Based on ARConv, ARNet replaces standard convolutions in a U-Net-style backbone to better handle scale-varying structures.

\textbf{BiMPan}~\cite{bimpan2023} proposes bidomain modeling that combines band-adaptive local modeling in the spatial domain with global detail reconstruction in the Fourier domain. It uses band-specific adaptive convolutions for local spectral structures and a Fourier global modeling module for global spatial details, and fuses the two to generate HRMS outputs.

\textbf{LAGConv}~\cite{lagconv2022} proposes a content-adaptive convolution that combines locally adaptive kernel weighting with a global harmonic bias. It predicts pixel-wise scaling factors over a shared base kernel and injects a global bias derived from pooled features to improve coherence. Plugging LAGConv into a lightweight residual fusion network yields improved adaptation to spatially varying patterns.

\textbf{FusionNet}~\cite{fusionnet2020} is a detail-injection CNN that predicts spatial details from the band-wise difference between replicated PAN and upsampled LRMS using a ResNet-style backbone. The predicted residual is added to the upsampled LRMS to obtain the fused HRMS image.

\textbf{PanNet}~\cite{pannet2017} trains in the high-pass domain to emphasize spatial detail learning. It feeds high-frequency PAN and upsampled LRMS into a ResNet to predict detail residuals, which are injected through a spectra-mapping skip connection to produce the final HRMS output and improve cross-satellite generalization.

\subsection{Evaluation Metrics}
\label{app:metrics_impl}

Under the reduced-resolution (RR) protocol, the pseudo ground-truth HRMS image is available, and we report full-reference metrics between the fused result and the reference.
Specifically, $Q_4$ or $Q_8$ measures multiband reconstruction quality, SAM evaluates spectral-angle distortion, ERGAS measures global radiometric distortion, and SCC assesses spatial-detail consistency.
Higher values are better for $Q_4/Q_8$ and SCC, while lower values are better for SAM and ERGAS.

Under the full-resolution (FR) protocol, no ground-truth HRMS image is available.
We therefore report no-reference QNR-based metrics.
$D_{\lambda}$ measures spectral distortion, $D_s$ measures spatial distortion, and HQNR summarizes the overall no-reference quality.
Lower values are better for $D_{\lambda}$ and $D_s$, while higher values are better for HQNR.

All compared methods use the same PAN and dataset-provided upsampled LRMS inputs.
For each sensor and protocol, metrics are computed using the same preprocessing, valid-region handling, and test split to ensure a fair comparison.

\section{Implementation Details}
\label{app:impl}

\subsection{Stage I: Single-Band VAE}
\label{app:impl:stage1}

Stage~I adapts the Stable Diffusion v1.5 VAE to remote-sensing HRMS patches and converts it into a single-band encoder--decoder shared across all spectral bands.
We only modify the first and last convolutional layers that interface with image channels, while keeping all intermediate layers unchanged.
Given a multispectral patch $X\in\mathbb{R}^{B\times H\times W}$, we fold the spectral dimension into the batch dimension so that each band $X^{(b)}\in\mathbb{R}^{1\times H\times W}$ becomes one sample and is processed by the same single-band VAE.

\subsubsection{Training objective}
For a single-band input $x=X^{(b)}\in\mathbb{R}^{1\times H\times W}$, the encoder defines
$q_\phi(z\mid x)=\mathcal{N}\!\left(\mu_\phi(x),\mathrm{diag}(\sigma_\phi^2(x))\right)$ and the decoder reconstructs $\hat x=D_\psi(z)$ with $z\sim q_\phi(z\mid x)$.
Stage~I minimizes a weighted reconstruction--regularization objective:
\begin{equation}
\begin{aligned}
\mathcal{L}_{\mathrm{vae}}
&=\lambda_{\mathrm{rec}}\,
\mathbb{E}_{z\sim q_\phi(z\mid x)}
\!\left[
\ell_{\mathrm{rec}}\big(x,D_\psi(z)\big)
\right] \\
&\quad
+\lambda_{\mathrm{kl}}\,
\mathrm{KL}\!\left(
q_\phi(z\mid x)\ \|\ \mathcal{N}(0,I)
\right).
\end{aligned}
\label{eq:app_stage1_vae_loss}
\end{equation}
where $\ell_{\mathrm{rec}}$ is a pixel-space reconstruction criterion.
We average Eq.~\eqref{eq:app_stage1_vae_loss} over bands and training patches by folding the band dimension into the batch dimension.

\subsubsection{Single-band conversion and weight initialization}
Let the original SD-VAE encoder input convolution be
\begin{equation}
\mathrm{Conv}_{\text{in}}(x)=W_{\text{in}} * x + b_{\text{in}},\qquad
W_{\text{in}}\in\mathbb{R}^{C\times 3\times k\times k},\ b_{\text{in}}\in\mathbb{R}^{C},
\end{equation}
and the converted single-band convolution be
\begin{equation}
\mathrm{Conv}'_{\text{in}}(g)=W'_{\text{in}} * g + b'_{\text{in}},\qquad
W'_{\text{in}}\in\mathbb{R}^{C\times 1\times k\times k}.
\end{equation}
We initialize $W'_{\text{in}}$ by enforcing functional equivalence on a pseudo-RGB embedding:
\begin{equation}
T(g)=
\begin{bmatrix}
\alpha_R g\\
\alpha_G g\\
\alpha_B g
\end{bmatrix},
\qquad
\alpha_R+\alpha_G+\alpha_B=1.
\label{eq:app_pseudo_rgb}
\end{equation}
Requiring $\mathrm{Conv}'_{\text{in}}(g)=\mathrm{Conv}_{\text{in}}(T(g))$ at initialization and using channel-wise linearity of convolution,
\begin{align}
\mathrm{Conv}_{\text{in}}(T(g))
&=\sum_{c\in\{R,G,B\}}
W_{\text{in}}^{(c)} * (\alpha_c g)
+ b_{\text{in}} \nonumber\\
&=
\Big(\sum_{c\in\{R,G,B\}}
\alpha_c W_{\text{in}}^{(c)}\Big) * g
+ b_{\text{in}}.
\end{align}
we obtain
\begin{equation}
W'_{\text{in}}=\sum_{c\in\{R,G,B\}} \alpha_c W_{\text{in}}^{(c)},
\qquad
b'_{\text{in}}=b_{\text{in}}.
\label{eq:app_conv_in_init}
\end{equation}
We use the standard luminance coefficients $(\alpha_R,\alpha_G,\alpha_B)=(0.299,0.587,0.114)$.

For the decoder, let
\begin{equation}
\begin{aligned}
y &= \mathrm{Conv}_{\text{out}}(h)
= W^{\mathrm{vae}}_{\text{out}} * h
+ b^{\mathrm{vae}}_{\text{out}},\\
W^{\mathrm{vae}}_{\text{out}}
&\in\mathbb{R}^{3\times C\times k\times k},\\
b^{\mathrm{vae}}_{\text{out}}
&\in\mathbb{R}^{3}.
\end{aligned}
\end{equation}
We define the single-band output as the luminance projection $\hat g=\sum_c \alpha_c y_c$.
By linearity,
\begin{equation}
\begin{aligned}
W'_{\text{out}}
&=\sum_{c\in\{R,G,B\}}
\alpha_c
\left(W^{\mathrm{vae}}_{\text{out}}\right)^{(c)},\\
b'_{\text{out}}
&=\sum_{c\in\{R,G,B\}}
\alpha_c
\left(b^{\mathrm{vae}}_{\text{out}}\right)^{(c)}.
\end{aligned}
\label{eq:app_conv_out_init}
\end{equation}
Eqs.~\eqref{eq:app_conv_in_init}--\eqref{eq:app_conv_out_init} ensure that, at initialization, the converted VAE behaves identically to the original SD-VAE applied to the embedding $T(g)$, reducing distribution drift after channel conversion.

\begin{figure*}[!t]
  \centering
  \includegraphics[width=\linewidth]{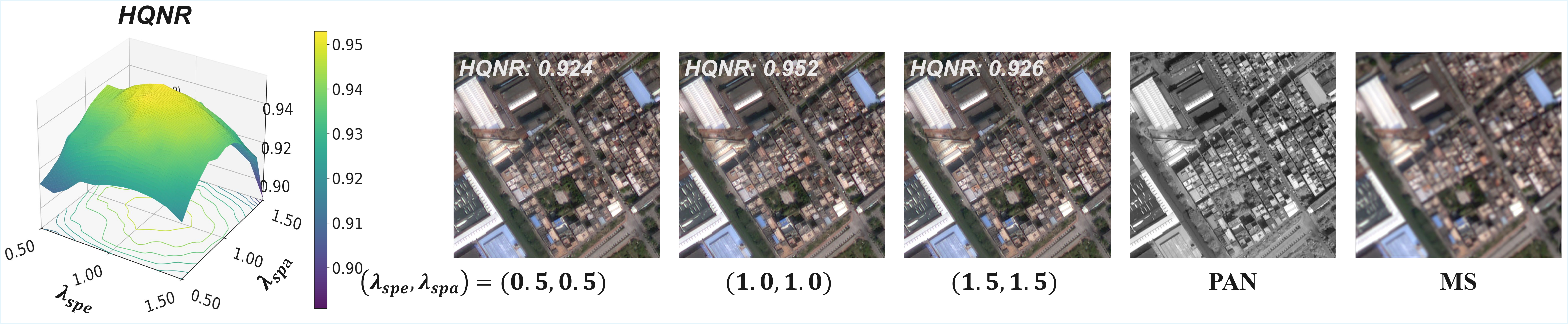}
  \caption{Test-time control by $(\lambda_{\mathrm{spe}},\lambda_{\mathrm{spa}})$.
  \textbf{Left:} HQNR response surface over $\lambda_{\mathrm{spe}},\lambda_{\mathrm{spa}}\in\{0.5,0.6,\ldots,1.5\}$, where each grid point is evaluated under an identical inference pipeline.
  \textbf{Right:} Pan-sharpened outputs on the same test sample for three representative settings $(0.5,0.5)$, $(1.0,1.0)$, and $(1.5,1.5)$, with their corresponding HQNR scores; the PAN and LRMS inputs are shown for reference.}
  \label{fig:lspe_lspa}
\end{figure*}

\subsubsection{Latent scaling factor}
After Stage~I converges, we calibrate a global latent scaling factor $\kappa$ and keep it fixed throughout Stage~II training and inference.
Stage~II performs diffusion in the VAE latent space under the convention that the clean latent has near-unit scale.
After single-band conversion and fine-tuning, the latent magnitude may drift, so we re-estimate a global scale $\kappa$ after Stage~I:
\begin{equation}
z_0^{(b)}=\kappa\,E\!\left(X^{(b)}\right),
\qquad
\hat X^{(b)}=D\!\left(\hat z_0^{(b)}/\kappa\right),
\label{eq:app_scale_apply}
\end{equation}
where $E(\cdot)$ and $D(\cdot)$ are the VAE encoder and decoder.

Since Stage~II uses the posterior mean as the clean latent, we set $E(x)=\mu_\phi(x)$ and calibrate $\kappa$ using
\begin{equation}
u:=\mu_\phi(x)\in\mathbb{R}^{d},
\end{equation}
where $d$ is the flattened latent dimensionality.

Approximating $u$ by a Gaussian with mean $m=\mathbb{E}[u]$ and covariance $\Sigma=\mathbb{E}[(u-m)(u-m)^\top]$, scaling $z=\kappa u$ yields
$z\sim\mathcal{N}(\kappa m,\kappa^2\Sigma)$.
We choose $\kappa>0$ to best match the unit-Gaussian convention:
\begin{equation}
\kappa^\star=\arg\min_{\kappa>0}\mathrm{KL}\!\left(\mathcal N(\kappa m,\kappa^2\Sigma)\ \|\ \mathcal N(0,I)\right).
\end{equation}
Using the Gaussian KL,
\begin{align}
\mathrm{KL}
&=\frac12\Big(\mathrm{tr}(\kappa^2\Sigma)+\|\kappa m\|_2^2-d-\log\det(\kappa^2\Sigma)\Big)\nonumber\\
&=\frac12\Big(\kappa^2(\mathrm{tr}\Sigma+\|m\|_2^2)-d-d\log\kappa^2\Big)+\text{const},
\end{align}
so $\partial\mathrm{KL}/\partial\kappa=0$ gives
\begin{equation}
\begin{aligned}
\kappa^{\star 2}
&=\frac{d}{\mathrm{tr}\Sigma+\|m\|_2^2}
=\frac{d}{\mathbb{E}\|u\|_2^2},\\
\kappa^\star
&=\sqrt{\frac{d}{\mathbb{E}\|u\|_2^2}}
=\frac{1}{
\sqrt{\mathbb{E}\!\left[
\tfrac{1}{d}\|u\|_2^2
\right]}}.
\end{aligned}
\label{eq:app_kappa_opt}
\end{equation}
Thus, $\kappa^\star$ makes the average latent energy $\mathbb{E}[\|z\|_2^2/d]$ close to $1$.

For DDPM-style forward diffusion $x_t=\sqrt{\bar\alpha_t}\,x_0+\sqrt{1-\bar\alpha_t}\,\epsilon$ with $\epsilon\sim\mathcal N(0,I)$,
a variance shift $\mathrm{Var}(x_0)\mapsto c^2\mathrm{Var}(x_0)$ changes the effective signal-to-noise ratio
\begin{equation}
\mathrm{SNR}_t=\frac{\bar\alpha_t\,\mathrm{Var}(x_0)}{1-\bar\alpha_t}
\ \mapsto\
\frac{\bar\alpha_t\,c^2\,\mathrm{Var}(x_0)}{1-\bar\alpha_t},
\end{equation}
misaligning the designed noise schedule.
Eq.~\eqref{eq:app_kappa_opt} enforces $\mathrm{Var}(x_0)\approx 1$ in the per-dimension energy sense, restoring schedule compatibility.

We approximate the expectation in Eq.~\eqref{eq:app_kappa_opt} with $N=10{,}000$ randomly sampled training patches.
For $u_i=\mu_\phi(x_i)\in\mathbb{R}^d$,
\begin{equation}
\widehat{s^2}
=\frac{1}{N}\sum_{i=1}^{N}\left(\frac{1}{d}\|u_i\|_2^2\right),
\qquad
\kappa=\frac{1}{\sqrt{\widehat{s^2}+\epsilon}},
\end{equation}
where $\epsilon$ is a small constant for stability.
We accumulate $\widehat{s^2}$ and then fix $\kappa$ for all Stage~II training and inference.

\subsection{Stage II: Latent Conditional Diffusion}
\label{app:impl:stage2}

\subsubsection{Training objective}
Stage~II trains the denoiser by noise prediction in the VAE latent space.
For each training tuple $(P,M,X,S)\sim\mathcal{D}$ and band index $b$, we form the clean latent
$z_0^{(b)}=\kappa_{\mathrm{vae}}E\!\left(X^{(b)}\right)$ and sample a timestep $t\sim\mathrm{Unif}\{1,\ldots,T\}$ with noise $\epsilon^{(b)}\sim\mathcal{N}(0,I)$:
\begin{equation}
z_t^{(b)}=\sqrt{\bar{\alpha}_t}\,z_0^{(b)}+\sqrt{1-\bar{\alpha}_t}\,\epsilon^{(b)}.
\label{eq:app_stage2_forward}
\end{equation}
The denoiser predicts $\hat{\epsilon}^{(b)}=f_\theta\!\left(z_t^{(b)},t;\,P,\tilde M^{(b)},E(s_{S,b})\right)$ and is trained with
\begin{equation}
\mathcal{L}_{\mathrm{diff}}
=\mathbb{E}_{(P,M,X,S)}\,
\mathbb{E}_{b,t,\epsilon^{(b)}}\Big[\big\|\epsilon^{(b)}-\hat{\epsilon}^{(b)}\big\|_2^2\Big].
\label{eq:app_stage2_diff_loss}
\end{equation}
In Stage~II, we keep the VAE and the main SD-UNet trunk frozen and optimize only the lightweight conditional modules that generate and inject residual controls.

\subsubsection{Dual conditional branches}
We implement spatial and spectral conditioning with two lightweight control branches attached to a largely frozen SD-UNet trunk.
Each branch follows the trunk multi-resolution layout and block indexing, but uses reduced channels and learns only residual controls.
We use a constant width ratio $\rho=\tfrac{1}{4}$ across all levels.
Let $c^\ell$ be the trunk channel width at UNet level $\ell$.
The branch width is $c_j^\ell=\rho\,c^\ell$ for $j\in\{\mathrm{spa},\mathrm{spe}\}$, with integer rounding when needed.

Let $h^\ell$ denote the trunk feature at level $\ell$.
For branch $j\in\{\mathrm{spa},\mathrm{spe}\}$, we compute a branch feature $s_j^\ell$ and inject a residual control $\Delta_j^\ell$ back into the trunk.
We adopt a hybrid coupling rule:
\begin{equation}
s_j^\ell=
\begin{cases}
\Phi_j^\ell\!\left(\Pi^\ell_{t\rightarrow j}(h^\ell),\ C_j\right), & \ell\in L_{\mathrm{enc}},\\[2pt]
\Phi_j^\ell\!\left(s_j^{\ell-1},\ C_j\right), & \ell\in L_{\mathrm{mid+dec}},
\end{cases}
\qquad
\Delta_j^\ell=\Pi^\ell_{j\rightarrow t}(s_j^\ell),
\label{eq:app_branch_hybrid_coupling}
\end{equation}
where $C_{\mathrm{spa}}=P$ and $C_{\mathrm{spe}}=\tilde M^{(b)}$.
In encoder blocks, the trunk feature $h^\ell$ is mapped to the branch space via $\Pi^\ell_{t\rightarrow j}$ to improve alignment.
In middle and decoder blocks, trunk-to-branch feedback is removed and the branches propagate only through $s_j^{\ell-1}$.
At all levels, the branch-to-trunk adapter $\Pi^\ell_{j\rightarrow t}$ produces the control residual $\Delta_j^\ell$.

\begin{figure*}[!t]
  \centering
  \includegraphics[width=\linewidth]{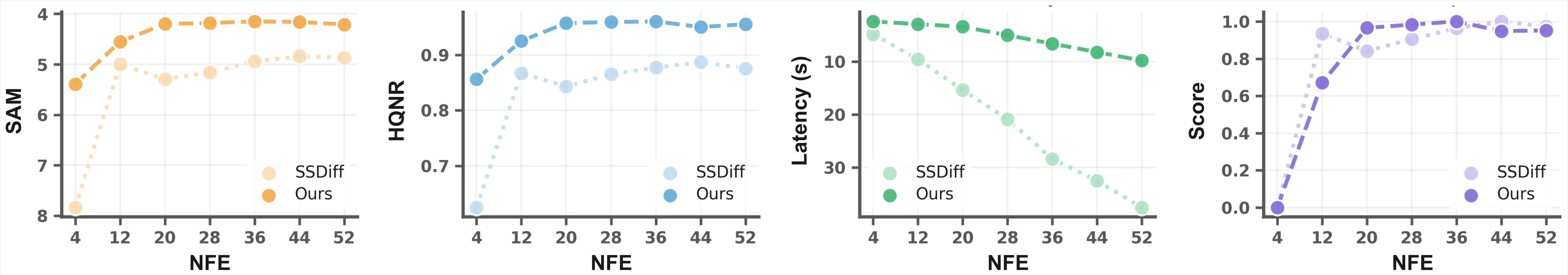}
  \caption{\textbf{Effect of the reverse-step budget.}
  We vary the number of reverse updates $K$ under a fixed inference configuration.
  From left to right: RR quality (SAM$\downarrow$), FR quality (HQNR$\uparrow$), wall-clock latency per reconstructed image, and an overall score.
  Our method reaches near-saturated quality at $\mathrm{NFE}\approx 20$ with substantially lower latency than the pixel-space diffusion baseline SSDiff.}
  \label{fig:K_sweep}
\end{figure*}

\begin{table*}[!t]
\centering
\caption{Sensor-specific parameters used for metadata prompt instantiation.}
\label{tab:prompt_params}
\small
\setlength{\tabcolsep}{5pt}
\renewcommand{\arraystretch}{1.22}

\begin{threeparttable}
\begin{adjustbox}{max width=\linewidth,center}
\begin{tabularx}{\linewidth}{@{}>{\bfseries}lccc>{\raggedright\arraybackslash}X@{}}
\toprule
\multirow{2}{*}{Sensor} &
\multicolumn{2}{c}{Ground Sampling Distance} &
\multirow{2}{*}{Bands} &
\multirow{2}{*}{MS band name and wavelength range (nm)} \\
\cmidrule(lr){2-3}
& PAN (m) & MS (m) & & \\
\midrule

GF2
& 1.00 & 4.00 & 4
& \textbf{Blue} [450,520]; \textbf{Green} [520,590];
  \textbf{Red} [630,690]; \textbf{NIR} [770,890] \\

QB
& 0.60 & 2.40 & 4
& \textbf{Blue} [450,520]; \textbf{Green} [520,600];
  \textbf{Red} [630,690]; \textbf{NIR} [760,900] \\

WV3
& 0.31 & 1.24 & 8
& \textbf{Coastal} [400,450]; \textbf{Blue} [450,510];
  \textbf{Green} [510,580]; \textbf{Yellow} [585,625];
  \textbf{Red} [630,690]; \textbf{RedEdge} [705,745];
  \textbf{NIR1} [770,895]; \textbf{NIR2} [860,1040] \\

WV2\tnote{a}
& 0.46 & 1.84 & 8
& \textbf{Coastal} [400,450]; \textbf{Blue} [450,510];
  \textbf{Green} [510,580]; \textbf{Yellow} [585,625];
  \textbf{Red} [630,690]; \textbf{RedEdge} [705,745];
  \textbf{NIR1} [770,895]; \textbf{NIR2} [860,1040] \\

\bottomrule
\end{tabularx}
\end{adjustbox}

\begin{tablenotes}[flushleft]
\footnotesize
\item[a] WV2 is used only for zero-shot evaluation and is not involved in training, validation, early stopping, or model selection.
\end{tablenotes}
\end{threeparttable}
\end{table*}

\subsubsection{Test-time control via \texorpdfstring{$(\lambda_{\mathrm{spe}},\lambda_{\mathrm{spa}})$}{lambda spe, lambda spa}}
We inject the spectral/spatial branch residuals into the trunk through a fixed frequency split:
\begin{equation}
h^{\ell}_{1}=h^{\ell}+\lambda_{\mathrm{spe}}\,L\!\left(\Delta^{\ell}_{\mathrm{spe}}\right), \qquad
h^{\ell}_{\mathrm{out}}=h^{\ell}_{1}+\lambda_{\mathrm{spa}}\,H\!\left(\Delta^{\ell}_{\mathrm{spa}}\right),
\end{equation}
where $L$ is a fixed low-pass operator and $H(\cdot)=I-L(\cdot)$ is the complementary high-pass operator.
We train with $\lambda_{\mathrm{spe}}=\lambda_{\mathrm{spa}}=1$ and tune them only at inference.

Let $\widehat{L}(\omega)$ denote the Fourier response of $L$.
Using $H=I-L$, the injected feature in the frequency domain becomes
\begin{equation}
\widehat{h^\ell_{\mathrm{out}}}(\omega)
=\widehat{h^\ell}(\omega)
+\lambda_{\mathrm{spe}}\widehat{L}(\omega)\widehat{\Delta^\ell_{\mathrm{spe}}}(\omega)
+\lambda_{\mathrm{spa}}\bigl(1-\widehat{L}(\omega)\bigr)\widehat{\Delta^\ell_{\mathrm{spa}}}(\omega),
\end{equation}
indicating that $\lambda_{\mathrm{spe}}$ controls the gain of the low-frequency components, whereas $\lambda_{\mathrm{spa}}$ controls the gain of the high-frequency components.
Moreover, if the split is approximately energy-orthogonal, i.e., $\langle L(x),H(y)\rangle \approx 0$, then
\begin{align}
&\bigl\|
\lambda_{\mathrm{spe}}L(\Delta_{\mathrm{spe}})
+\lambda_{\mathrm{spa}}H(\Delta_{\mathrm{spa}})
\bigr\|_2^2 \nonumber\\
&=
\lambda_{\mathrm{spe}}^2
\|L(\Delta_{\mathrm{spe}})\|_2^2
+
\lambda_{\mathrm{spa}}^2
\|H(\Delta_{\mathrm{spa}})\|_2^2 \nonumber\\
&\quad
+2\lambda_{\mathrm{spe}}\lambda_{\mathrm{spa}}
\left\langle
L(\Delta_{\mathrm{spe}}),
H(\Delta_{\mathrm{spa}})
\right\rangle \nonumber\\
&\approx
\lambda_{\mathrm{spe}}^2
\|L(\Delta_{\mathrm{spe}})\|_2^2
+
\lambda_{\mathrm{spa}}^2
\|H(\Delta_{\mathrm{spa}})\|_2^2 .
\end{align}
which explains the stable two-knob trade-off under a fixed inference pipeline.

We sweep
\begin{equation}
\mathcal{G}=\Bigl\{\lambda \,\big|\, \lambda\in\{0.5,\,0.6,\,\ldots,\,1.5\}\Bigr\},
\end{equation}
with a step size of $0.1$, while keeping model weights, preprocessing, frequency split, sampler, noise schedule, the number of sampling steps, and random seed fixed.
For a given test sample, we select the best setting by
\begin{equation}
(\lambda_{\mathrm{spe}}^{\star},\lambda_{\mathrm{spa}}^{\star})
=\arg\max_{\lambda_{\mathrm{spe}},\lambda_{\mathrm{spa}}\in\mathcal{G}}\;\mathrm{HQNR}(\lambda_{\mathrm{spe}},\lambda_{\mathrm{spa}}).
\end{equation}
For the sample shown in Fig.~\ref{fig:lspe_lspa}, the optimum is $(\lambda_{\mathrm{spe}}^{\star},\lambda_{\mathrm{spa}}^{\star})=(1.0,\,0.9)$.

\subsubsection{Inference step budget}
Given PAN $P\!\in\!\mathbb{R}^{1\times H\times W}$ and LRMS $M\!\in\!\mathbb{R}^{B\times h\times w}$ with scale ratio $r=H/h=W/w$,
we first align LRMS to the PAN scale via bicubic upsampling:
\begin{equation}
\tilde M=\mathrm{Up}(M)\in\mathbb{R}^{B\times H\times W}.
\end{equation}
We then perform band-wise latent conditional diffusion.
For each band $b$, we sample $z_T^{(b)}\!\sim\!\mathcal{N}(0,I)$ and run a UniPC reverse process under the SD1.5 discrete noise schedule:
\begin{equation}
\begin{aligned}
z_{t_{k-1}}^{(b)}
&=\mathrm{UniPC}\!\Big(
z_{t_k}^{(b)},
\hat\epsilon_\theta\!\big(
z_{t_k}^{(b)},t_k;\,P,\tilde M^{(b)}
\big),
t_k\!\rightarrow\! t_{k-1}
\Big),\\
&\qquad\qquad k=K,\dots,1 .
\end{aligned}
\end{equation}
where each reverse update uses one denoiser forward pass, hence $\mathrm{NFE}=K$.

Because inference is stochastic, independently sampling $\{z_T^{(b)}\}_{b=1}^B$ can introduce band-wise sampling drift.
We therefore draw a single latent per image and broadcast it across bands:
\begin{equation}
z_T\sim\mathcal{N}(0,I),\qquad z_T^{(b)}=z_T,\ \forall b,
\end{equation}
which removes initial cross-band discrepancy.
For a fair step-budget comparison, we fix the random seed, and thus the same $z_T$, across different $K$.

Fig.~\ref{fig:K_sweep} shows the quality--speed trade-off under an identical inference pipeline.
As $\mathrm{NFE}$ increases, quality improves quickly at small budgets and becomes near-saturated around $\mathrm{NFE}\approx 20$ for our method, while wall-clock latency scales approximately linearly with $\mathrm{NFE}$:
\begin{equation}
\mathrm{Latency}(K)\approx \tau_0+\tau\cdot \mathrm{NFE}=\tau_0+\tau\cdot K.
\end{equation}
Therefore, we adopt the smallest budget in the saturation regime, with default $K=20$, to balance reconstruction quality and runtime.

\begin{figure*}[!t]
    \centering
    \includegraphics[width=\linewidth]{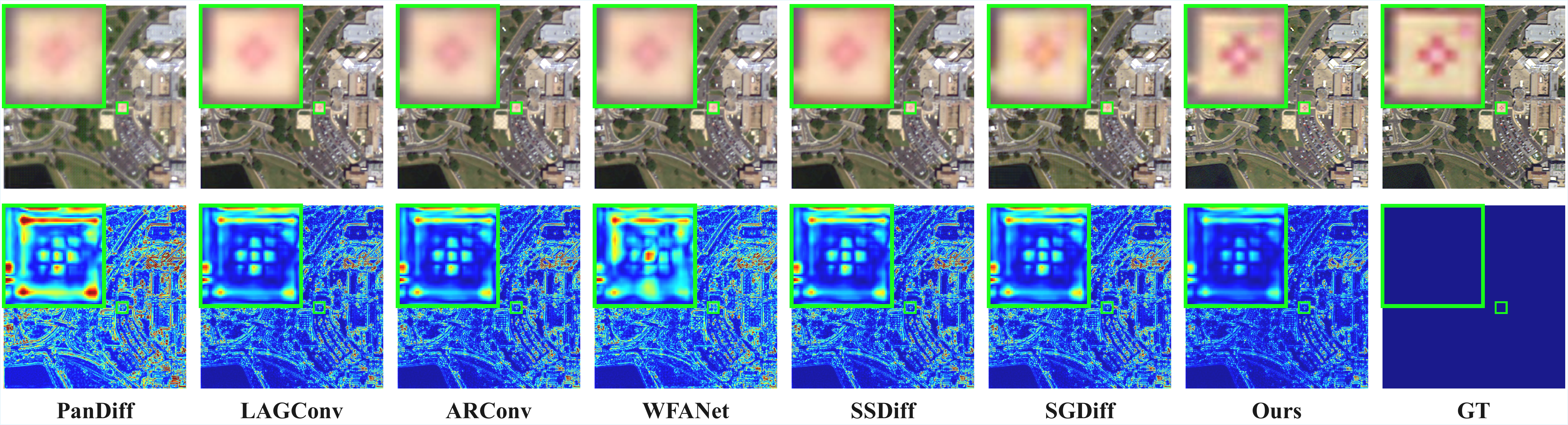}
    \caption{Visual comparison on the WorldView-2 dataset at reduced resolution.}
    \label{fig:wv2_rr_vis}
\end{figure*}

\begin{figure*}[!t]
    \centering
    \includegraphics[width=\linewidth]{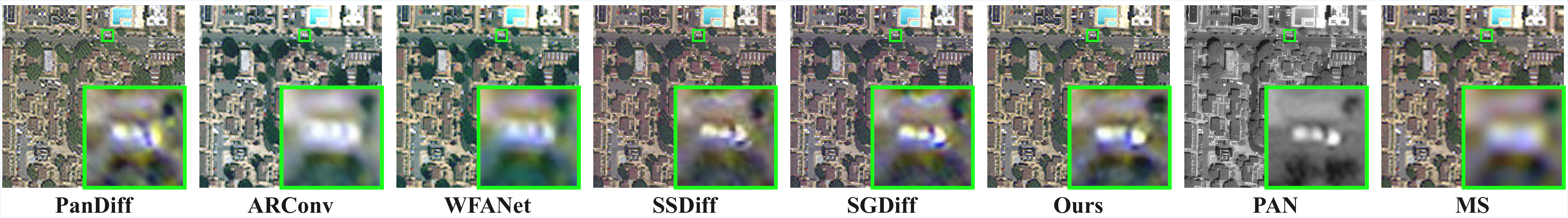}
    \caption{Visual comparison on the WorldView-2 dataset at full resolution.}
    \label{fig:wv2_fr_vis}
\end{figure*}

\subsubsection{Prompt templates}
We condition the denoiser on sensor-specific and band-specific metadata via text prompts encoded by a frozen CLIP text encoder.
For each sensor $S$ and multispectral band $b$, we construct the prompt by concatenating two components:
\begin{equation}
s_{S,b} = s^{\mathrm{data}}_{S}\ \Vert\ s^{\mathrm{band}}_{S,b},
\end{equation}
where $\Vert$ denotes string concatenation with a single whitespace separator.
The first component $s^{\mathrm{data}}_{S}$ describes the sensor configuration, while the second component $s^{\mathrm{band}}_{S,b}$ specifies the attributes of band $b$.
The band ordering follows the channel order in the dataset.

We use two unified templates that work across all sensors:
\begin{center}
\setlength{\fboxsep}{4pt}
\fbox{
\begin{minipage}{0.92\linewidth}
\small\ttfamily
Sensor <S>. PAN GSD <g\_pan> m. MS GSD <g\_ms> m. MS bands <B>.\\
Band <nu>. Wavelength [<lambda\_lo>,<lambda\_hi>] nm.
\end{minipage}}
\end{center}
The first line defines $s^{\mathrm{data}}_{S}$, and the second line defines $s^{\mathrm{band}}_{S,b}$.
Here, $g^{\mathrm{pan}}_{S}$ and $g^{\mathrm{ms}}_{S}$ denote the ground sampling distances for PAN and MS bands, respectively;
$B_{S}$ is the total number of MS bands; $\nu_{S,b}$ is the human-readable band name; and the wavelength range is specified by the lower and upper bounds $\lambda^{\mathrm{lo}}_{S,b}$ and $\lambda^{\mathrm{hi}}_{S,b}$ in nanometers.
The text encoder produces an embedding $E(s_{S,b})$, which is injected into the diffusion U-Net via cross-attention layers.
We use the same templates for both training and inference to ensure consistency.

For sensor WV3 and its fourth MS band, Yellow, the instantiated prompt is:
\begin{center}
\setlength{\fboxsep}{4pt}
\fbox{
\begin{minipage}{0.92\linewidth}
\small\ttfamily
Sensor WV3. PAN GSD 0.31 m. MS GSD 1.24 m. MS bands 8.\\
Band Yellow. Wavelength [585,625] nm.
\end{minipage}}
\end{center}
Equivalently,
\begin{equation}
s_{\mathrm{WV3},4}
=
s^{\mathrm{data}}_{\mathrm{WV3}}
\ \Vert\
s^{\mathrm{band}}_{\mathrm{WV3},4}.
\end{equation}

\section{Additional Zero-Shot Visual Results}
\label{app:quant_vis}

This appendix complements the main paper with additional qualitative comparisons that are not included due to page limits.
Unless otherwise stated, we follow the dataset and evaluation protocols described in Appendix~\ref{app:datasets}.
The examples focus on WorldView-2, which is used only for zero-shot testing and is never used for training, validation, early stopping, or model selection.

\subsection{WorldView-2 Zero-Shot Transfer}
\label{app:transfer:zeroshot}

We additionally present zero-shot transfer results on WorldView-2, abbreviated as WV2.
The model is trained on other sensors and directly evaluated on WV2 with no retraining.
This transfer setting is challenging because sensor-dependent spectral responses and PAN-to-MS coupling vary across platforms, and mismatched priors often lead to unstable colors or over-confident sharpening artifacts.

Figures~\ref{fig:wv2_rr_vis} and~\ref{fig:wv2_fr_vis} provide representative results under RR and FR.
In Fig.~\ref{fig:wv2_rr_vis}, the zoom-in focuses on a compact high-contrast pattern that tends to collapse into a blurry blob when fine details are not properly recovered.
CC-Pan restores a clearer localized configuration in the zoomed region and yields a noticeably weaker error response, suggesting improved reconstruction of fine structures under cross-sensor transfer.

In Fig.~\ref{fig:wv2_fr_vis}, some baselines exhibit visible chromatic artifacts near the bright structure in the zoom-in, with unnatural tints and smeared boundaries.
CC-Pan follows the PAN-indicated geometry more faithfully while keeping the local tone consistent with the MS input, leading to more natural textures without exaggerated color shifts.
Together, the WV2 visualizations support that the learned cross-sensor prior generalizes well to an unseen sensor domain.

\end{document}